\documentclass[10pt,twocolumn,letterpaper]{article}

\usepackage{iccv}
\usepackage{times}
\usepackage{epsfig}
\usepackage{graphicx}
\usepackage{amsmath}
\usepackage{amssymb}

\usepackage{pifont}
\usepackage{booktabs}
\usepackage{multirow}
\usepackage{capt-of}
\usepackage{makecell}
\usepackage{colortbl}
\usepackage{dsfont}
\usepackage{comment}
\usepackage[dvipsnames]{xcolor}

\usepackage[accsupp]{axessibility}  %

\usepackage{tocloft}

\makeatletter
\renewcommand{\numberline}[1]{%
  \@cftbsnum #1\@cftasnum\hspace*{1em}\@cftasnumb%
}
\makeatother

\usepackage[pagebackref=true,breaklinks=true,colorlinks,bookmarks=false]{hyperref}

\usepackage[capitalize]{cleveref}
\crefname{section}{Sec.}{Secs.}
\Crefname{section}{Section}{Sections}
\Crefname{table}{Table}{Tables}
\crefname{table}{Tab.}{Tabs.}

\newcommand{\tablestyle}[2]{\setlength{\tabcolsep}{#1}\renewcommand{\arraystretch}{#2}\centering\footnotesize}
\newcommand{\tablestylesmaller}[2]{\setlength{\tabcolsep}{#1}\renewcommand{\arraystretch}{#2}\centering\scriptsize}
\renewcommand{\paragraph}[1]{\vspace{1.25mm}\noindent\textbf{#1}}

\newcommand{\cmark}{\ding{51}}%
\newcommand{\xmark}{\ding{55}}%

\definecolor{baselinecolor}{gray}{.9}
\definecolor{darkgreen}{rgb}{0.13, 0.55, 0.13}

\let\originalleft\left
\let\originalright\right
\renewcommand{\left}{\mathopen{}\mathclose\bgroup\originalleft}
\renewcommand{\right}{\aftergroup\egroup\originalright}

\iccvfinalcopy %

\def\iccvPaperID{5097} %

\def\OURS{SimGCD}

\begin{document}

\title{Parametric Classification for Generalized Category Discovery: A Baseline Study}

\author{%
  Xin Wen\textsuperscript{$1$}\footnotemark[1] \qquad
  Bingchen Zhao\textsuperscript{$2$}\footnotemark[1] \qquad
  Xiaojuan Qi\textsuperscript{$1$} \\
  \textsuperscript{$1$}The University of Hong Kong \quad
  \textsuperscript{$2$}University of Edinburgh \\
  \tt\small \{wenxin,xjqi\}@eee.hku.hk \qquad
  \tt bingchen.zhao@ed.ac.uk \\
}

\twocolumn[{%
\maketitle
\renewcommand\twocolumn[1][]{#1}%
\maketitle
\begin{center}
    \centering
    \includegraphics[width=\textwidth]{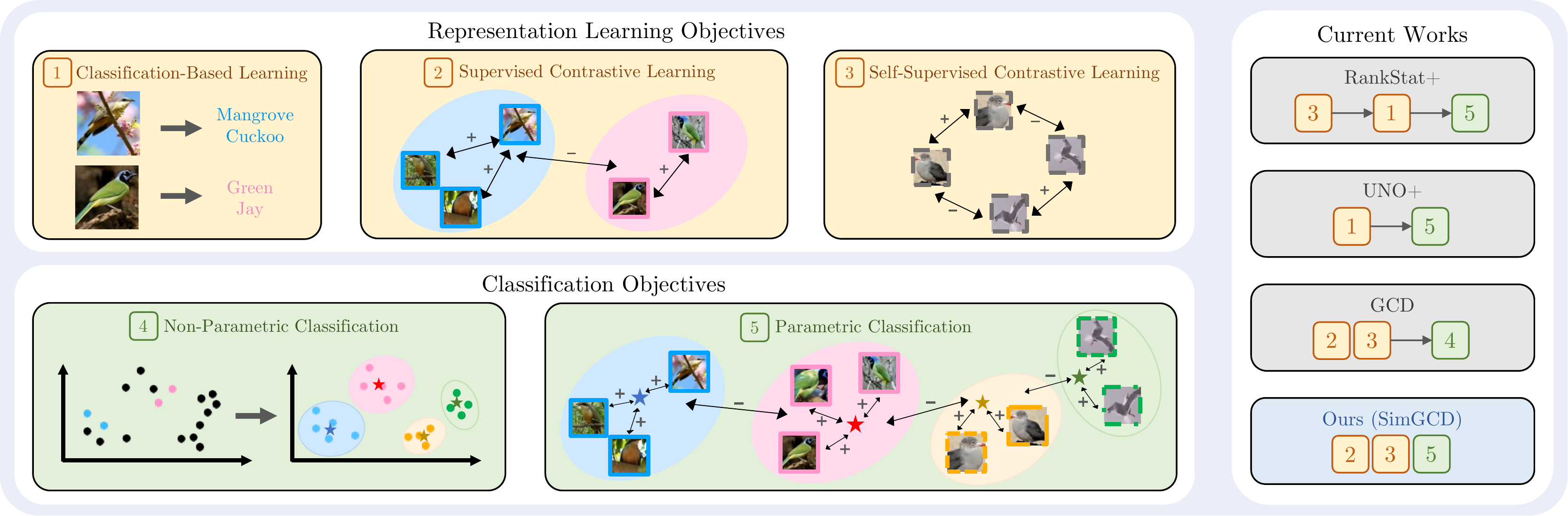}
    \captionof{figure}{
    Left: building blocks for representation learning or classifier learning;
    Right: overall abstraction of current works, where `$\rightarrow$' separates different stages of the method.
    Our work builds on GCD~\cite{vaze22generalized}, and jointly trains a parametric classifier.
    } \label{fig:teaser}
    \vspace{1.5em}
\end{center}%
}]

{\renewcommand{\thefootnote}{\fnsymbol{footnote}}
\footnotetext[1]{Equal contribution.}}

\maketitle

\addtocontents{toc}{\protect\setcounter{tocdepth}{0}}

\begin{abstract}
\vspace{-.5em}

Generalized Category Discovery (GCD) aims to discover novel categories in unlabelled datasets using knowledge learned from labelled samples.
Previous studies argued that parametric classifiers are prone to overfitting to seen categories, and endorsed using a non-parametric classifier formed with semi-supervised $k$-means.
However, in this study, we investigate the failure of parametric classifiers, verify the effectiveness of previous design choices when high-quality supervision is available, and identify unreliable pseudo-labels as a key problem. We demonstrate that two prediction biases exist: the classifier tends to predict seen classes more often, and produces an imbalanced distribution across seen and novel categories. 
Based on these findings, we propose a simple yet effective parametric classification method that benefits from entropy regularisation, achieves state-of-the-art performance on multiple GCD benchmarks and shows strong robustness to unknown class numbers.
We hope the investigation and proposed simple framework can serve as a strong baseline to facilitate future studies in this field.
Our code is available at: \url{https://github.com/CVMI-Lab/SimGCD}.

\end{abstract}

\vspace{-1.4em}
\section{Introduction}\label{sec:intro}

With large-scale labelled datasets, deep learning methods can surpass humans in recognising images~\cite{resnet}. %
However, it is not always possible to collect large-scale human annotations for training deep learning models. %
Therefore, there is a rich body of recognition models that focus on learning with a large number of unlabelled data.
Among them, semi-supervised learning~(SSL)~\cite{oliver2018realistic,berthelot2019mixmatch,sohn2020fixmatch} is regarded as a promising approach, yet with the assumption that labelled instances are provided for each of the categories the model needs to classify.
Generalized category discovery~(GCD)~\cite{vaze22generalized} is recently formalised to relax this assumption by assuming the unlabelled data can also contain similar yet distinct categories from the labelled data.
The goal of GCD is to learn a model that is able to classify the already-seen categories in the labelled data, and more importantly, \textit{jointly discover the new categories in the unlabelled data and make correct classifications}.
Developing a strong method for this problem could help us better utilise the easily available large-scale unlabelled datasets.

Previous works~\cite{vaze22generalized,han21autonovel,fini2021unified,cao21orca} approach this problem from two perspectives:
learning generic feature representations to facilitate the discovery of novel categories, and generating pseudo clusters/labels for unlabelled data to guide the learning of a classifier.
The former is often achieved by using self-supervised learning methods~\cite{han21autonovel,zhao21novel,gidaris2018unsupervised_rotnet,he2019moco,caron2021emerging,dvp} to improve the generalization ability of features to novel categories. 
For constructing the classifier, earlier works~\cite{han21autonovel,zhao21novel,zhong2021openmix,cao21orca,fini2021unified} adopt a parametric approach that builds a learnable classifier on top of the extracted features. The classifier is jointly optimised with the backbone using labelled data and pseudo-labelled data.

\begin{figure}[t]
\centering
\includegraphics[width=\linewidth]{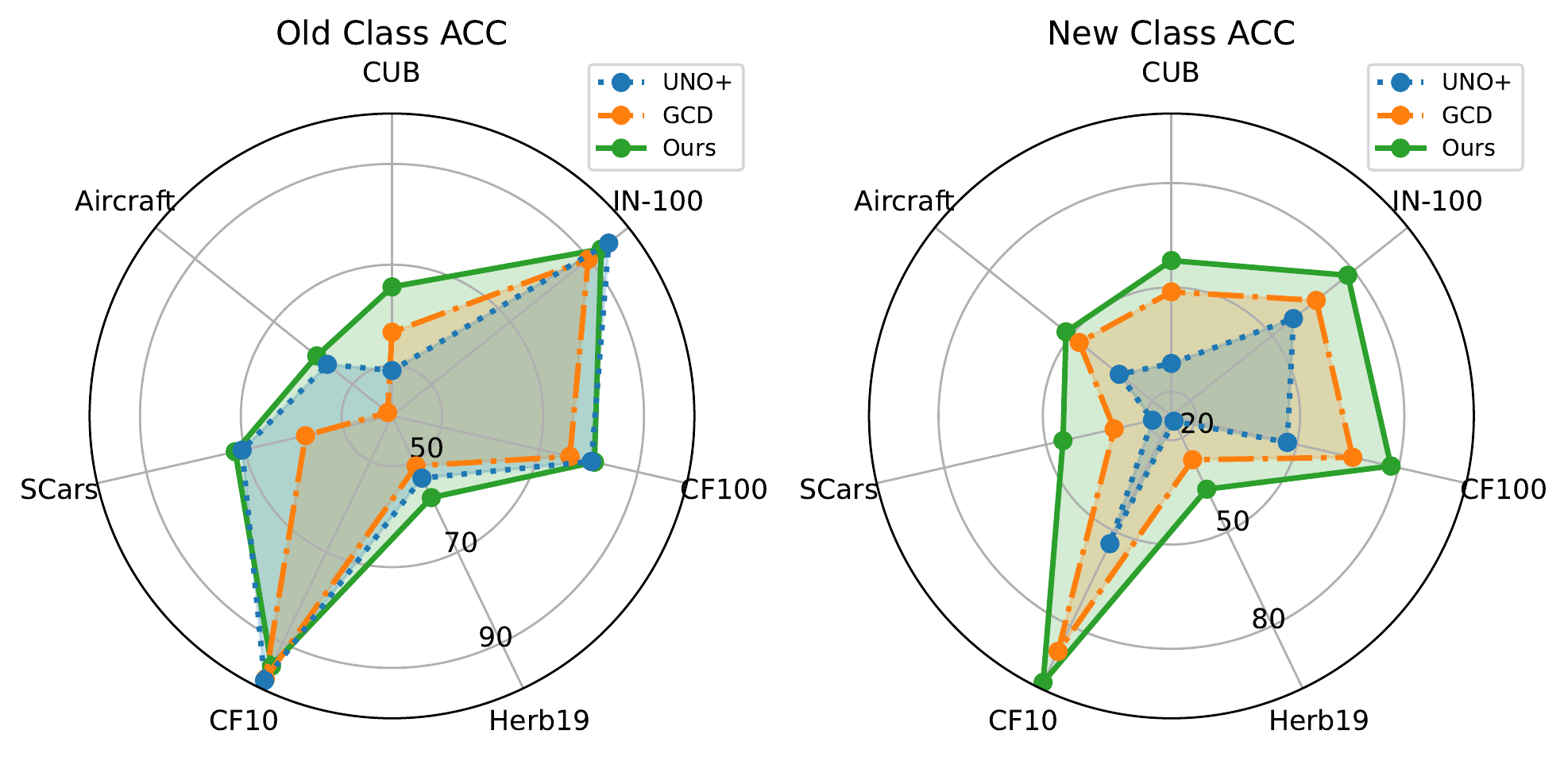}
\caption{
    \textbf{Performance overview.}
    Prior parametric classification method (UNO+~\cite{fini2021unified}) shows highly degraded performance in `New' classes. The non-parametric classification work (GCD~\cite{vaze22generalized}) performs better, but at the sacrifice of `Old' class and high inference cost. Our method shows that parametric classification can work well on both metrics.
    }
\label{fig:radar}
\end{figure}

However, recent research shows~\cite{vaze22generalized,fei2022xcon} that parametric classifiers are prone to overfit to seen categories (see \cref{fig:radar}) and thus promote using a non-parametric classifier such as $k$-means clustering. 
Albeit obtaining promising results, the non-parametric classifiers suffer from heavy computation costs on large-scale datasets due to quadratic complexity of the clustering algorithm. 
Besides, unlike a learnable parametric classifier, the non-parametric method loses the ability to jointly optimise the separating hyperplane of all categories in a learnable manner, potentially being sub-optimal.

This motivates us to revisit the reason that makes previous parametric classifiers fail to recognise novel classes.
In a series of investigations (\cref{sec:pilot}) from the view of supervision quality, we verify the effectiveness of prior design choices in feature representations and training paradigms when strong supervision is available, and conclude that the key to previous parametric classifiers’ degraded performance is unreliable pseudo labels.
By diagnosing the statistics of its predictions, we identify severe prediction biases within the model, \ie, the bias towards predicting more `Old' classes than `New' classes (\cref{fig:errors}) and the bias of producing imbalanced pseudo-labels across all classes (\cref{fig:longtail}).

Based on these findings, we thus present a simple parametric classification baseline for generalized category discovery (see \cref{fig:teaser,fig:framework}).
The representation learning objective follows GCD~\cite{vaze22generalized}, and the classification objective is simply cross-entropy for labelled samples and self-distillation~\cite{caron2021emerging,assran2022masked} for unlabelled samples. 
Besides, an entropy regularisation term is also adopted to overcome biased predictions by enforcing the model to predict more uniformly distributed labels across all possible categories. 
Empirically, we indeed observe that our method produces more balanced pseudo-labels (\cref{fig:ours_error,fig:ours_longtail}) and achieves a large performance gain on multiple GCD benchmarks (\cref{subtab:ssb,subtab:generic,tab:herb19}), indicating that the two types of biases we identified are the core reason why the parametric-classifier-based approach performs poorly for GCD.
Additionally, we observe that the entropy regulariser could also be used to enforce robustness towards an unknown number of categories (\cref{fig:ks,fig:largeK_analysis}), this could further ease the deployment of parametric classifiers for GCD in real-world scenarios.

Our contributions are summarised as follows: 
(1) We revisit the design choices of parametric classification and conclude the key factors that make it fail for GCD.
(2) Based on the analysis, we propose a simple yet effective parametric classification method.
(3) Our method achieves SOTA on multiple popular GCD benchmarks, challenging the recent promotion of non-parametric classification for this task.

\section{Related Works}

\paragraph{Semi-Supervised Learning}
(SSL) has been an important research topic where a number of methods have been proposed~\cite{berthelot2019mixmatch,sohn2020fixmatch,tarvainen2017mean}. 
SSL assumes that the labelled instances are available for all possible categories in the unlabelled dataset; 
the objective is to learn a model to perform classification using both the labelled samples as well as the large-scale available unlabelled data.
One of the most effective methods for SSL is the consistency-based method, where the model is forced to learn consistent representations of two different augmentations of the same image~\cite{sohn2020fixmatch,berthelot2019mixmatch,tarvainen2017mean}.
Furthermore, it is also shown that self-supervised representation learning is helpful for the task of SSL~\cite{zhai2019s4l,rebuffi2020semi} as it can provide a strong representation for the task.

\paragraph{Open-Set Semi-Supervised Learning}
considers the case where the unlabelled data may contain outlier data points that do not belong to any of the categories in the labelled training set.
The goal is to learn a classifier for the labelled categories from a noisy unlabelled dataset~\cite{mtc_opensetssl,saito2021openmatch,opensetssl1,opensetssl2}.
As this problem only focuses on the performance of the labelled categories, the outlier from novel categories are simply rejected and no further classification is needed.

\paragraph{Generalized Category Discovery}
(GCD) is a relatively new problem recently formalised in Vaze~\etal~\cite{vaze22generalized}, and is also studied in a parallel line of work termed open-world semi-supervised learning~\cite{cao21orca,sun2023opencon}. Different from the common assumption of SSL~\cite{oliver2018realistic}, GCD does not assume the unlabelled dataset comes from the same class set as the labelled dataset, posing a greater challenge for designing an effective model.
GCD can be seen as a natural extension of the novel category discovery (NCD) problem~\cite{Han2019learning} where it is assumed that the unlabelled dataset and the labelled dataset do not have any class overlap, thus baselines for NCD~\cite{han21autonovel,zhao21novel,zhong2021openmix,zhong2021neighborhood,fini2021unified} can be adopted for the GCD problem by extending the classification head to have more outputs~\cite{vaze22generalized}.
The incremental setting of GCD is also explored~\cite{zhao2023incremental,incd2022}.
It is pointed out in~\cite{vaze22generalized} that a non-parametric classifier formed using semi-supervised $k$-means can outperform strong parametric classification baselines from NCD~\cite{han21autonovel,fini2021unified} because it can alleviate the overfitting to seen categories in the labelled set.
In this paper, we revisit this claim and show that parametric classifiers can reach stronger performance than non-parametric classifiers.%

\paragraph{Deep Clustering}
aims at learning a set of semantic prototypes from unlabelled images with deep neural networks. 
Considering that no label information is available, the focus is on how to obtain reliable pseudo-labels. 
While early attempts rely on hard labels produced by $k$-means~\cite{caron2018deep}, there has been a shift towards soft labels produced by optimal transport~\cite{ym2019selflabel,caron2020unsupervised}, and more recently sharpened predictions from an exponential moving average-updated teacher model~\cite{caron2021emerging,assran2022masked}.
Deep clustering has shown strong potential for unsupervised representation learning~\cite{caron2018deep,ym2019selflabel,caron2020unsupervised,caron2021emerging,assran2022masked}, unsupervised semantic segmentation~\cite{cho2021picie,wen2022slotcon}, semi-supervised learning~\cite{assran2021semi}, and novel category discovery~\cite{fini2021unified}.
In this work, we study the techniques that make strong parametric classifiers for GCD with inspirations from deep clustering.

\section{On the Failure of Parametric Classification} \label{sec:pilot}

In order to explore the reason that makes previous parametric classifiers fail to recognise `New' classes for generalized category discovery, this section presents preliminary studies to reveal the role of two major components: representation learning (\cref{subsec:whichrep}) and pseudo-label quality on unseen classes (\cref{subsec:decouple}).
These have led to \textit{conflicting choices} of previous works, but why?
We show a unified viewpoint (\cref{fig:clusterpos,fig:decouple}), and emphasise that \textit{taking pseudo-label quality into account} is important for selecting the suitable design choice.
This then led to our diagnosis of what makes the degenerated pseudo-labels (\cref{subsec:bias_pred}), and motivated our de-biased pseudo-labelling strategy.

\subsection{Investigation Setting}

\paragraph{Generalized category discovery.}
Given an unlabelled dataset $\mathcal{D}^u=\left\{(\boldsymbol{x}_i^u, {y}_i^u)\right\}\in \mathcal{X}\times \mathcal{Y}_u$ where $\mathcal{Y}_u$ is the label space of the unlabelled samples, the goal of GCD is to learn a model to categorise the samples in $\mathcal{D}^u$ using the knowledge from a labelled dataset $\mathcal{D}^l=\left\{(\boldsymbol{x}_i^l, {y}_i^l)\right\}\in \mathcal{X}\times \mathcal{Y}_l$ where $\mathcal{Y}_l$ is the label space of labelled samples and $\mathcal{Y}_l \subset \mathcal{Y}_u$.
We denote the number of categories in $\mathcal{Y}_u$ as $K_u=|\mathcal{Y}_u|$, 
it is common to assume the number of categories is known \textit{a-priori}~\cite{han21autonovel,zhao21novel,zhong2021openmix,fini2021unified}, or can be estimated using off-the-shelf methods~\cite{Han2019learning,vaze22generalized}.

\paragraph{Representation learning.}
For representation learning, we follow GCD~\cite{vaze22generalized}, which applies supervised contrastive learning~\cite{khosla2020supervised} on labelled samples, and self-supervised contrastive learning~\cite{chen2020simple} on all samples (detailed in \cref{subsec:method_rep}).

\paragraph{Classifier.}
We follow UNO~\cite{fini2021unified} to adopt a \textit{prototypical classifier}. Take $f(\boldsymbol{x})$ as the feature vector of an image $\boldsymbol{x}$ extracted using from the backbone $f$, the procedure for producing logits is $\boldsymbol{l} = \frac{1}{\tau}(\boldsymbol{w} / ||\boldsymbol{w}||)^{\top} (f(\boldsymbol{x}) / ||f(\boldsymbol{x})||)$. Here $\tau$ is the temperature value that scales up the norm of $\boldsymbol{l}$ and facilitates optimisation of the cross-entropy loss~\cite{wang2017normface}. 

\paragraph{Training settings.}
We train with varying supervision qualities. The \textit{minimal supervision} setting utilises only the labels in $\mathcal{D}^l$, while the \textit{oracle supervision} setting assumes all samples are labelled (both $\mathcal{D}^l$ and $\mathcal{D}^u$).
Besides, we study two practical settings that adopt pseudo labels for unlabelled samples in $\mathcal{D}^u$: \textit{self-label} that predicts pseudo-labels with the Sinkhorn Knopp algorithm following~\cite{fini2021unified}, and \textit{self-distil}, which depicts another pseudo-labelling strategy as in \cref{fig:framework} and will be introduced in detail in \cref{subsec:method_cls}.
For all settings, we only employ a cross-entropy loss on the (pseudo-)labelled samples on hand for classification.
Note that unless otherwise stated, \textit{this is done on decoupled features, thus representation learning is unaffected}.

\subsection{Which Representation to Build Your Classifier?} \label{subsec:whichrep}

\paragraph{Motivation.}
Following the trend of deep clustering that focuses on self-supervised representation learning~\cite{caron2020unsupervised}, previous parametric classification work UNO~\cite{fini2021unified} fed the classifier with representations taken from the \textit{projector}. While in GCD~\cite{vaze22generalized}, significantly stronger performance is achieved with a non-parametric classifier built upon representations taken from the \textit{backbone}. We revisit this choice as follows.

\paragraph{Setting.}
Consider $f$ as the feature backbone, and $g$ is a multi-layer perceptron (MLP) projection head.
Given an input image $\boldsymbol{x}_i$, the representation from the \textit{backbone} can be written as $f(\boldsymbol{x}_i)$, and that from the \textit{projector} is $g(f(\boldsymbol{x}_i))$. 

\begin{figure}[b]
\centering
\includegraphics[width=\linewidth]{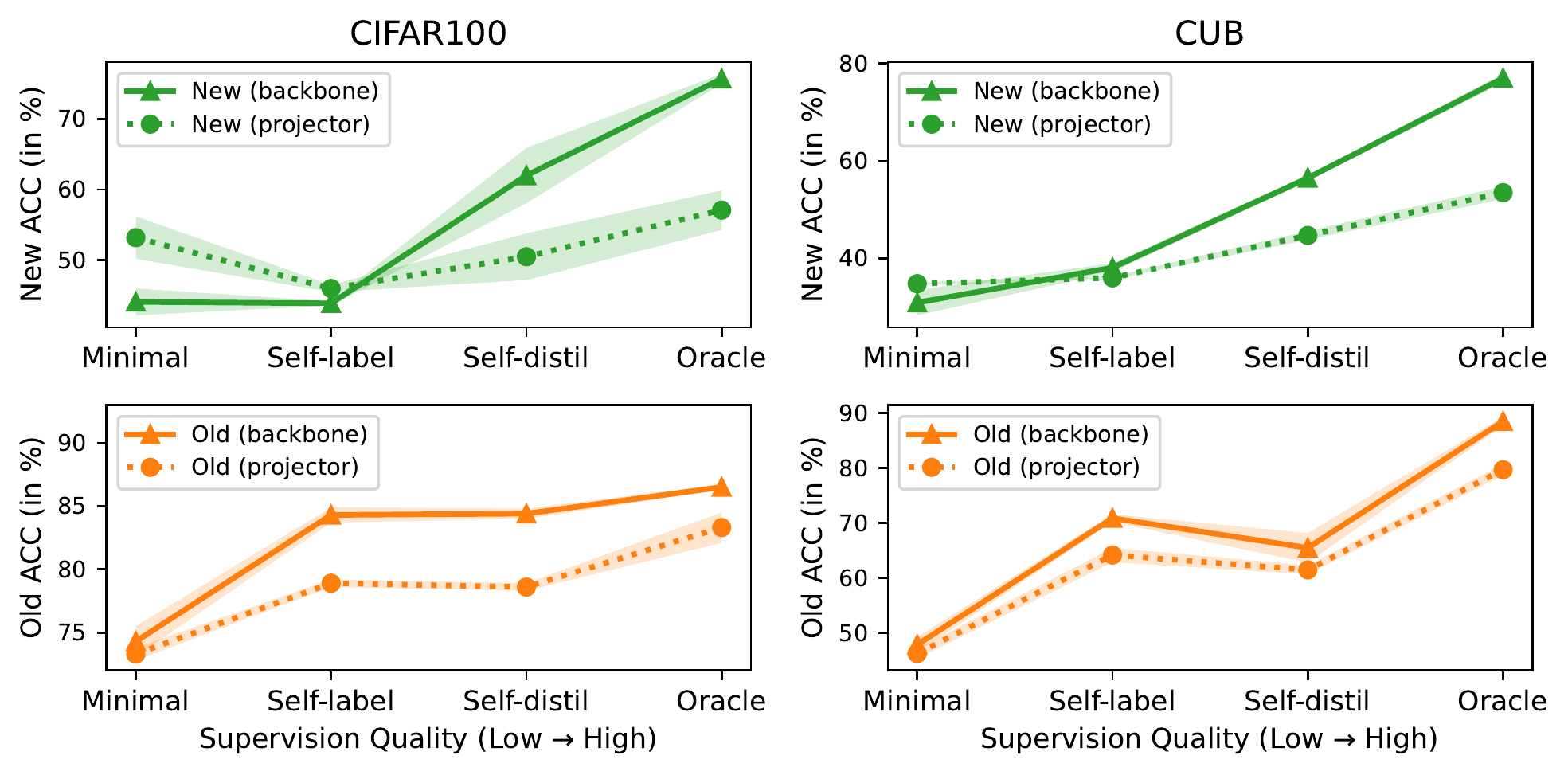}
\vspace{-1.5em}
\caption{
    \textbf{Results with different representations.}
    We build the classifier on \textit{post-backbone} or \textit{post-projector} representations, and train with varying supervision quality.
    Results on `Old' class consistently benefit from the post-backbone representations regardless of the supervision quality, while unleashing its potential on `New' class requires stronger pseudo labels.
}
\label{fig:clusterpos}
\end{figure}

\paragraph{Result \& discussion.}
As in \cref{fig:clusterpos}, the post-backbone feature space has a clearly higher upper bound for learning prototypical classifiers than the post-projector feature space.
Using a projector in self-supervised learning lets the projector focus on solving pretext tasks and allows the backbone to keep as much information as possible (which facilitates downstream tasks)~\cite{cui2022discriminability}.
But when good classification performance is all you need, our results suggest that the classification objective should build on post-backbone representations directly. The features post the projector might focus more on solving the pretext task and not be necessarily useful for the classification objective.
Note that high-quality pseudo labels are necessary to unleash the post-backbone representations' potential to recognise novel categories.

\subsection{Decoupled or Joint Representation Learning?} \label{subsec:decouple}

\paragraph{Motivation.}
Previous parametric classification methods, \eg, UNO~\cite{fini2021unified}, commonly tune the representations jointly with the classification objective.
On the contrary, in the two-stage non-parametric method GCD~\cite{vaze22generalized} where the performance in `New' classes is notably higher, classification/clustering is fully decoupled from representation learning, and the representations can be viewed as unaltered by classification.
In this part, we study 
whether the \textit{joint} learning strategy contributes to previous parametric classifiers' degraded performance in recognising `New' classes.

\paragraph{Setting.}
Consider $f(\boldsymbol{x})$ as the representation fed to the classifier, \textit{decoupled} training, as the previous settings adopted, indicates $f(\boldsymbol{x})$ is decoupled when computing the logits $\boldsymbol{l}$, thus the classification objective won't supervise representation learning. While for \textit{joint} training, the representations are jointly optimised by classification.

\begin{figure}[b]
\centering
\includegraphics[width=\linewidth]{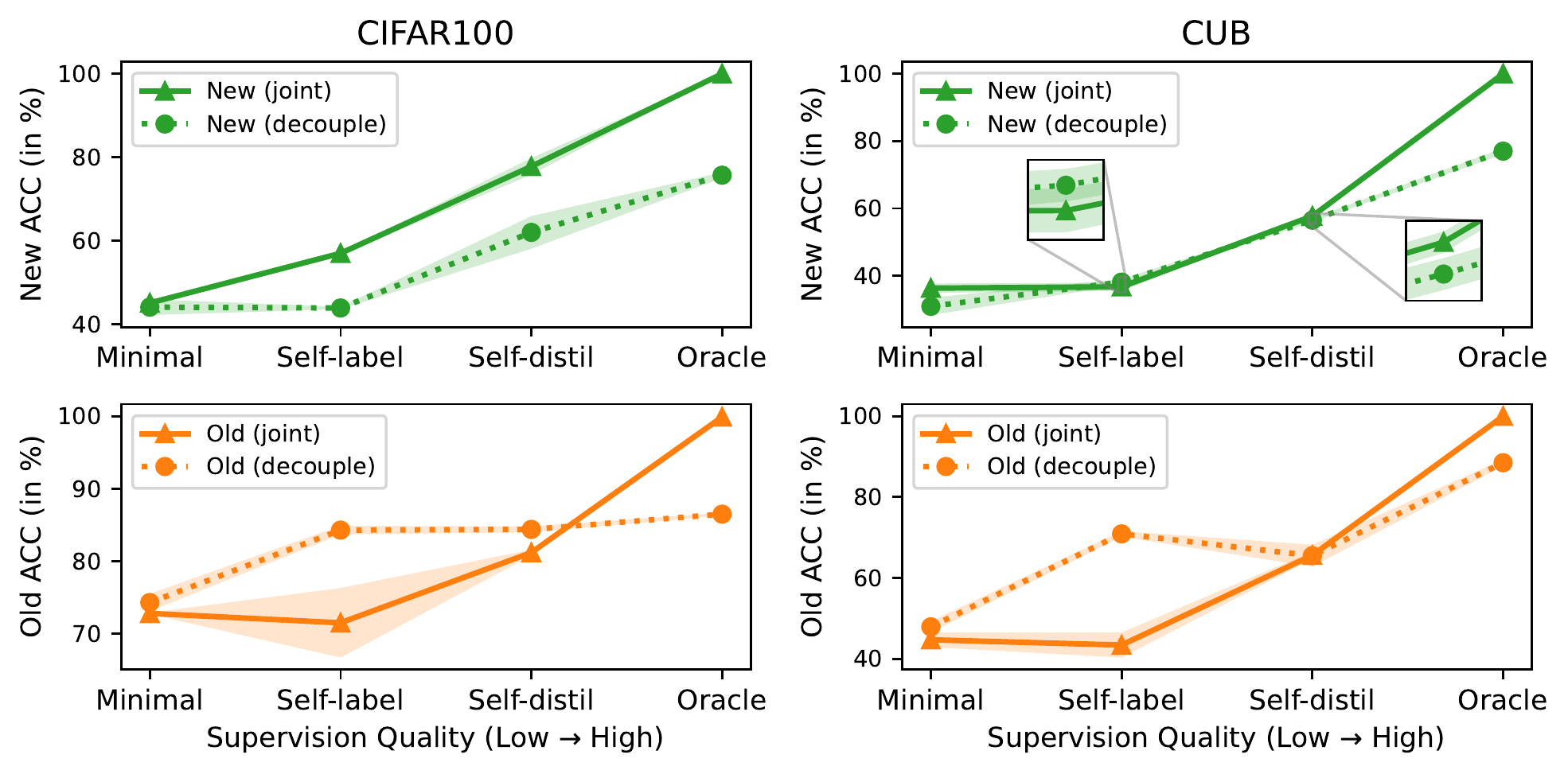}
\caption{
    \textbf{Results with different training paradigms.}
    \textit{Decouple} denotes the classifier adopts decoupled features, while \textit{joint} indicates the classification objective can affect representation learning. Joint training is helpful when high-quality supervision is available, otherwise, it could lead to degraded representations.
}
\label{fig:decouple}
\end{figure}

\paragraph{Result \& discussion.}
The results are illustrated in \cref{fig:decouple}.
When adopting the self-labelling strategy, there is a sharp drop in `Old' class performance on both datasets, while for the `New' classes, it can improve by 13 points on CIFAR100, and drop by a small margin on CUB.
In contrast, when a stronger pseudo-labelling strategy (self-distillation) or even oracle labels are utilised, we observe consistent gains from joint training.
This means that \textit{the joint training strategy does not necessarily result in UNO~\cite{fini2021unified}'s low performance in `New' classes}; on the contrary, it can even boost `New' class performance by a notable margin.
Our overall explanation is that UNO's framework could not make reliable pseudo-labels, thus restricting its ability to benefit from joint training. %
The joint training strategy is not to blame and is, in fact, helpful. When switching to a more advanced pseudo-labelling paradigm that produces higher-quality pseudo-labels, the help from joint training can be even more significant.

\subsection{The Devil Is in the Biased Predictions} \label{subsec:bias_pred}

\paragraph{Motivation.}
In \cref{subsec:whichrep,subsec:decouple}, we verified the effectiveness of two design choices when high-quality pseudo labels are available, and concluded the key to previous work's degraded performance is unreliable pseudo labels.
We then further diagnose the statistics of its predictions as follows.

\paragraph{Setting.}
We categorise the model's errors into four types: ``True Old'', ``False New'', ``False Old'', and ``True New'' according to the relationship between predicted and ground-truth class. \Eg, ``True New'' refers to predicting a `New' class sample to another `New' class, while ``False Old'' indicates predicting a `New' class sample as some `Old' class.

\begin{figure}[b]
\centering
\includegraphics[width=\linewidth]{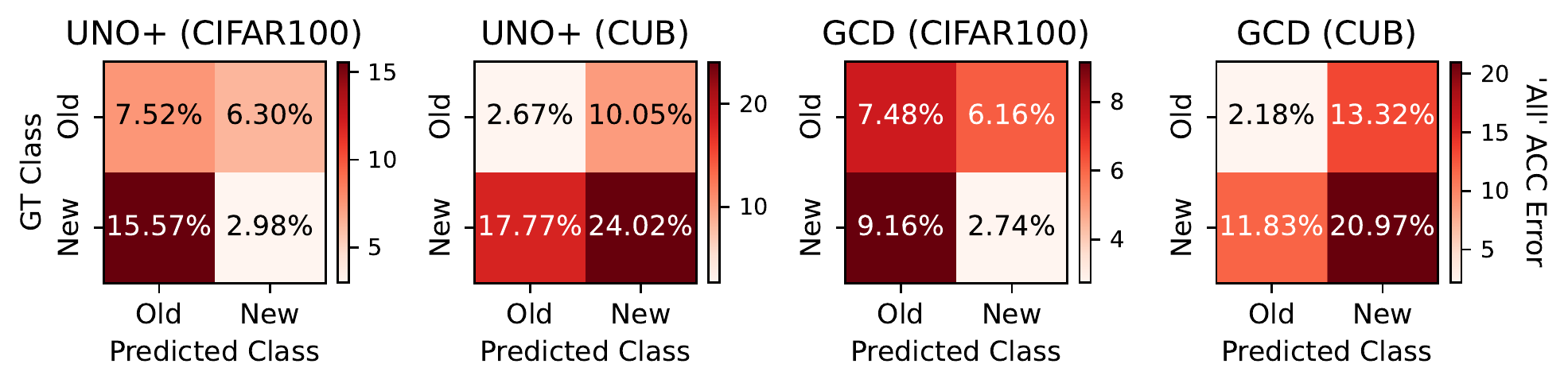}
\caption{
    \textbf{Prediction bias between `Old'/`New' classes.}
    We simplify the results to binary classification and categorise errors in `All' ACC into four types. Both works, especially UNO+, are prone to make ``False Old'' predictions, and many samples corresponding to `New' classes are misclassified as an `Old' class.
}
\label{fig:errors}
\end{figure}
\begin{figure}[b]
\centering
\includegraphics[width=\linewidth]{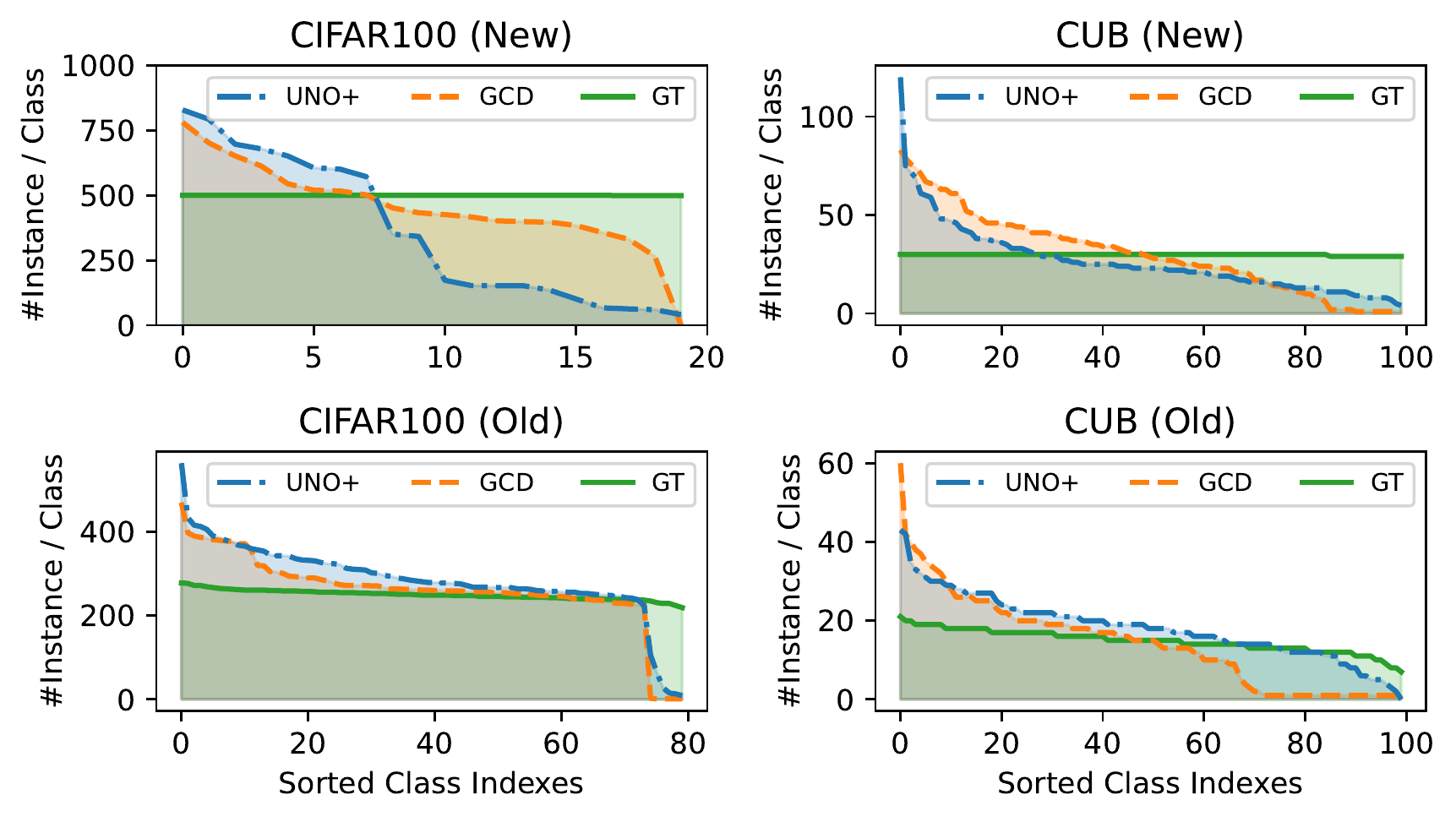}
\caption{
    \textbf{Prediction bias across `Old'/`New' classes.}
    We show the per-class prediction distributions. Both works, especially UNO+, are prone to make biased predictions. Across all classes, the predictions are unexpectedly biased towards the head classes.
}
\label{fig:longtail}
\end{figure}

\paragraph{Result \& discussion.}
We observe two types of prediction bias.
In \cref{fig:errors}, both works, especially UNO+~\cite{fini2021unified}, are prone to make ``False Old'' predictions. In other words, their predictions are biased towards `Old' classes. Besides, the ``True New'' errors are also notable, indicating that misclassification within `New' classes is also common.
We then depict the predictions' overall distribution across `Old'/`New' classes in \cref{fig:longtail}, and both works show highly biased predictions.
This double-bias phenomenon then motivated the prediction entropy regularisation design in our method.

\section{Method} \label{sec:method}
\begin{figure}[t]
\centering
\includegraphics[width=\linewidth]{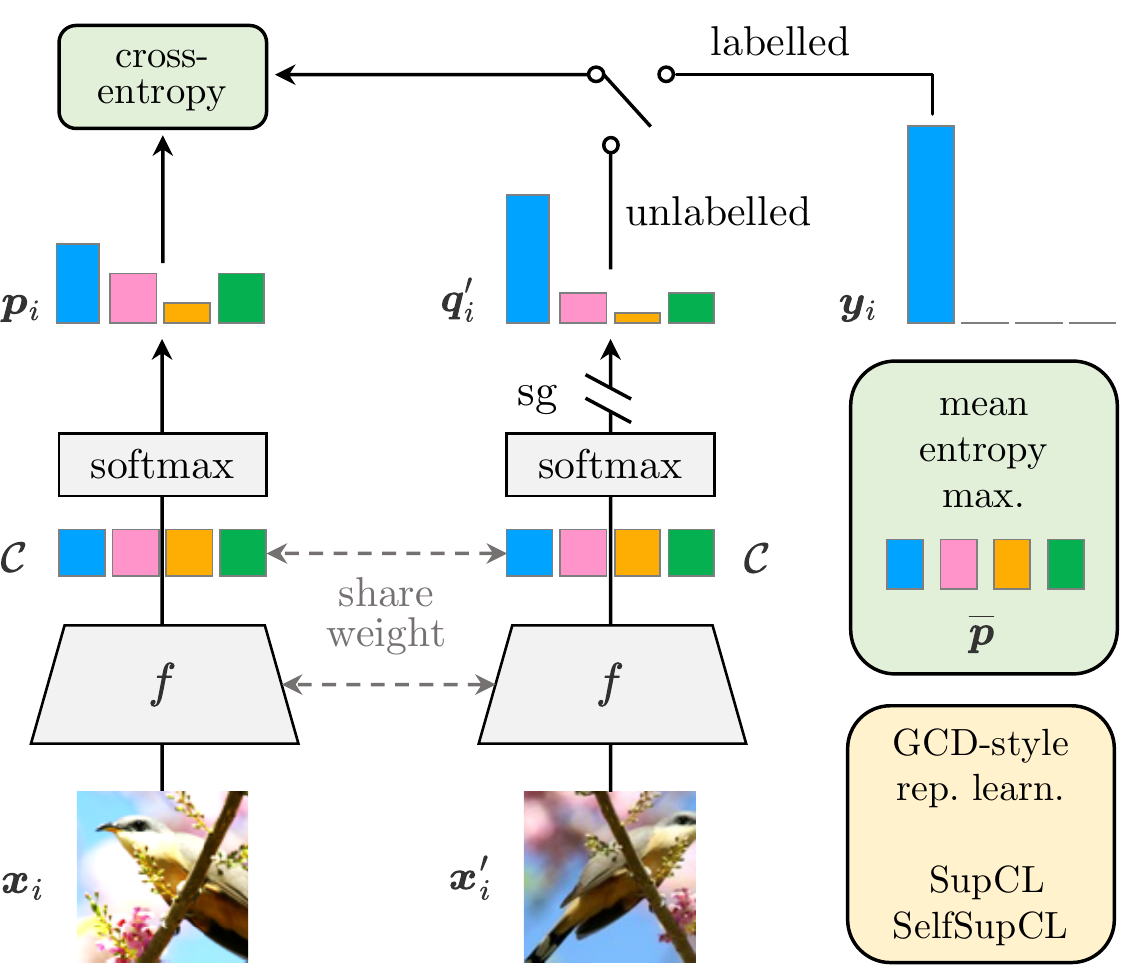}
\caption{
    \textbf{The overall framework of our method.}
    For unlabelled samples, the pseudo-labels are from sharpened predictions of another random augmented view. And for labelled samples, we simply adopt the ground truth. Details for representation learning and the mean-entropy-maximisation regulariser are omitted for simplicity, and please refer to the text. (Also see \cref{fig:teaser} for a high-level comparison with previous works)
}
\label{fig:framework}
\end{figure}

In this section, we present the whole picture of this simple yet effective method (see \cref{fig:framework}), a one-stage framework that builds on GCD~\cite{vaze22generalized}, and jointly trains a parametric classifier with self-distillation and entropy regularisation.
And in \cref{subsec:ablation}, we discuss the step-by-step changes that lead a simple baseline to our solution.
\subsection{Representation Learning} \label{subsec:method_rep}
Our representation learning objective follows GCD~\cite{vaze22generalized}, which is supervised contrastive learning~\cite{khosla2020supervised} on labelled samples, and self-supervised contrastive learning~\cite{chen2020simple} on all samples.
Formally, given two views (random augmentations) $\boldsymbol{x}_i$, and $\boldsymbol{x}_i^{\prime}$ of the same image in a mini-batch $B$, the self-supervised contrastive loss is written as:
\begin{equation}
\mathcal{L}^u_\text{rep}=\frac{1}{|B|} \sum_{i \in B}-\log \frac{\exp \left(\boldsymbol{z}_i^\top \boldsymbol{z}_i^{\prime} / \tau_u\right)}{\sum_i^{i \neq n} \exp \left(\boldsymbol{z}_i^\top \boldsymbol{z}_n^{\prime} / \tau_u\right)} \,,
\end{equation}
where the feature $\boldsymbol{z}_i=g\left(f\left(\boldsymbol{x}_i\right)\right)$ and is $\ell_2$-normalised, $f, g$ denote the backbone and the projection head, and $\tau_u$ is a temperature value.
The supervised contrastive loss is similar, and the major difference is that positive samples are matched by their labels, formally written as:
\begin{equation}
\mathcal{L}^s_\text{rep}=\frac{1}{|B^l|} \sum_{i \in B^l}\frac{1}{|\mathcal{N}_i|} \sum_{q \in \mathcal{N}_i} -\log \frac{\exp \left(\boldsymbol{z}_i^\top \boldsymbol{z}_q^{\prime} / \tau_c\right)}{\sum_i^{i \neq n} \exp \left(\boldsymbol{z}_i^\top \boldsymbol{z}_n^{\prime} / \tau_c\right)} \,,
\end{equation}
where $\mathcal{N}_i$ indexes all other images in the same batch that hold the same label as $\boldsymbol{x}_i$.
The overall representation learning loss is balanced with $\lambda$:
 $\mathcal{L}_\text{rep} = (1 - \lambda) \mathcal{L}^u_\text{rep} + \lambda \mathcal{L}^s_\text{rep}$,
where $B^{l}$ corresponds to the labelled subset of $B$.

\subsection{Parametric Classification} \label{subsec:method_cls}
Our parametric classification paradigm follows the self-distillation~\cite{caron2021emerging,assran2022masked} fashion.
Formally, with $K=|\mathcal{Y}_l \cup \mathcal{Y}_u|$ denoting the total number of categories, we randomly initialise a set of prototypes $\mathcal{C}=\{\boldsymbol{c}_1, \dots, \boldsymbol{c}_K\}$, each standing for one category.
During training, we calculate the soft label for each augmented view $\boldsymbol{x}_i$ by softmax on cosine similarity between the hidden feature $\boldsymbol{h}_i=f(\boldsymbol{x}_i)$ and the prototypes $\mathcal{C}$ scaled by $1/\tau_s$:
\begin{equation}
\boldsymbol{p}_{i}^{(k)}=\frac{\exp \left(\frac{1}{\tau_s}(\boldsymbol{h}_i/||\boldsymbol{h}_i||_2)^\top (\boldsymbol{c}_k / ||\boldsymbol{c}_k||_2)\right)}{\sum_{k^\prime} \exp \left(\frac{1}{\tau_s}(\boldsymbol{h}_i / ||\boldsymbol{h}_i||_2)^\top (\boldsymbol{c}_{k^\prime} / ||\boldsymbol{c}_{k^\prime}||_2)\right)} \,,
\end{equation}
and the soft pseudo-label $\boldsymbol{q}_i^\prime$ is produced by another view $\boldsymbol{x}_i$ with a sharper temperature $\tau_t$ in a similar fashion.
The classification objectives are then simply cross-entropy loss $\ell(\boldsymbol{q}^{\prime}, \boldsymbol{p}) = -\sum_{k} {\boldsymbol{q}^{\prime(k)}}\log{\boldsymbol{p}}^{(k)}$ between the predictions and pseudo-labels or ground-truth labels:
\begin{equation} \label{eq:selfdistill}
    \mathcal{L}^u_\text{cls} = \frac{1}{|B|}\sum_{i \in B}\ell(\boldsymbol{q}_i^\prime, \boldsymbol{p}_i) - \varepsilon H(\overline{\boldsymbol{p}}) ,
     \mathcal{L}^s_\text{cls} = \frac{1}{|B^l|}\sum_{i \in B^l}\ell(\boldsymbol{y}_i, \boldsymbol{p}_i)
     ,
\end{equation}
where $\boldsymbol{y}_i$ denote the one-hot label of $\boldsymbol{x}_i$.
We also adopt a mean-entropy maximisation regulariser~\cite{assran2022masked} for the unsupervised objective. 
Here $\overline{\boldsymbol{p}} = \frac{1}{2|B|}\sum_{i \in B}\left( \boldsymbol{p}_i+\boldsymbol{p}_i^\prime \right)$ denotes the mean prediction of a batch, and the entropy $H(\overline{\boldsymbol{p}}) = -\sum_k\overline{\boldsymbol{p}}^{(k)}\log\overline{\boldsymbol{p}}^{(k)}$.
Then the classification objective is $\mathcal{L}_\text{cls} = (1 - \lambda) \mathcal{L}^u_\text{cls} + \lambda \mathcal{L}^s_\text{cls}$, 
and the overall objective is simply $\mathcal{L}_\text{rep} + \mathcal{L}_\text{cls}$.

\paragraph{Discussions.}
Please note that this work doesn't aim to promote new methods but to examine existing solutions, provide insights into their failures and build a simple yet strong baseline solution.
The paradigm of producing pseudo-labels from sharpened predictions of another augmented view appears to resemble consistency-based methods~\cite{sohn2020fixmatch,berthelot2019mixmatch,tarvainen2017mean} in the SSL community.
However, despite differences in augmentation strategies and soft/hard pseudo-labels, our approach jointly performs category discovery and self-training style learning, while the SSL methods purely focus on bootstrapping itself with unlabelled data, and does not discover novel categories.
Besides, entropy regularisation is also explored in deep clustering to avoid trivial solution~\cite{assran2022masked}. In contrast, our method shows its help in overcoming the prediction bias between and within seen/novel classes~(\cref{fig:ours_error,fig:ours_longtail}), and enforcing robustness to unknown numbers of categories~(\cref{fig:ks}).

\section{Experiments}
\subsection{Experimental Setup}
\paragraph{Datasets.}
We validate the effectiveness of our method on the generic image recognition benchmark (including CIFAR10/100~\cite{cifar} and ImageNet-100~\cite{deng2009imagenet}), the recently proposed Semantic Shift Benchmark~\cite{vaze22openset} (SSB, including CUB~\cite{cub200}, Stanford Cars~\cite{stanfordcars}, and FGVC-Aircraft~\cite{aircraft}), and the harder Herbarium~19~\cite{tan2019herbarium} and ImageNet-1K~\cite{deng2009imagenet}.
For each dataset, we follow~\cite{vaze22generalized} to sample a subset of all classes as the labelled (`Old') classes $\mathcal{Y}_l$;
50\% of the images from these labelled classes are used to construct $\mathcal{D}^l$, and the remaining images are regarded as the unlabelled data $\mathcal{D}^u$. %
See~\cref{tab:datasplit} for statistics of the datasets we evaluate on.

\begin{table}[htbp]   
    \vspace{-.3em}
    \begin{center}
    \tablestyle{4pt}{1}
    \begin{tabular}{lcrrrr}
    \toprule
            &    & \multicolumn{2}{c}{Labelled}  & \multicolumn{2}{c}{Unlabelled}\\
                \cmidrule(rl){3-4}\cmidrule(rl){5-6}
    Dataset         & Balance   & \#Image   & \#Class   & \#Image   & \#Class \\
    \midrule
    CIFAR10~\cite{cifar}         & \cmark    & 12.5K     & 5         & 37.5K     & 10 \\
    CIFAR100~\cite{cifar}        & \cmark    & 20.0K     & 80        & 30.0K     & 100 \\
    ImageNet-100~\cite{deng2009imagenet}    & \cmark    & 31.9K     & 50        & 95.3K     & 100 \\
    CUB~\cite{cub200}             & \cmark    & 1.5K      & 100       & 4.5K      & 200 \\
    Stanford Cars~\cite{stanfordcars}   & \cmark    & 2.0K      & 98        & 6.1K      & 196 \\    
    FGVC-Aircraft~\cite{aircraft}   & \cmark    & 1.7K      & 50        & 5.0K      & 100 \\
    Herbarium 19~\cite{tan2019herbarium}    & \xmark    & 8.9K      & 341       & 25.4K     & 683 \\
    ImageNet-1K~\cite{deng2009imagenet}    & \cmark    & 321K     & 500        & 960K     & 1000 \\
    \bottomrule
    \end{tabular}
    \end{center}
    \vspace{-.8em}
    \caption{Statistics of the datasets we evaluate on.}\label{tab:datasplit}
    \vspace{-.9em}
\end{table}

\paragraph{Evaluation protocol.}
We evaluate the model performance with clustering accuracy (ACC) following standard practice~\cite{vaze22generalized}. 
During evaluation, given the ground truth $y^*$ and the predicted labels $\hat{y}$, the ACC is calculated as $\text{ACC} = \frac{1}{M} \sum_{i=1}^{M} \mathds{1}(y^*_i = p(\hat{y}_i))$ where $M = |\mathcal{D}^u|$, and $p$ is the optimal permutation that matches the predicted cluster assignments to the ground truth class labels.

\paragraph{Implementation details.}
Following GCD~\cite{vaze22generalized}, we train all methods with a ViT-B/16 backbone~\cite{dosovitskiy2020vit} pre-trained with DINO~\cite{caron2021emerging}.
We use the output of \texttt{[CLS]} token with a dimension of 768 as the feature for an image, and only fine-tune the last block of the backbone. %
We train with a batch size of 128 for 200 epochs with an initial learning rate of 0.1 decayed with a cosine schedule on each dataset.
Aligning with~\cite{vaze22generalized}, the balancing factor $\lambda$ is set to 0.35, and the temperature values $\tau_u$, $\tau_c$ as 0.07, 1.0, respectively.
For the classification objective, we set $\tau_s$ to 0.1, and $\tau_t$ is initialised to 0.07, then warmed up to 0.04 with a cosine schedule in the starting 30 epochs. %
All experiments are done with an NVIDIA GeForce RTX 3090 GPU.%

\subsection{Comparison With the State of the Arts}

\begin{table}[b]
\begin{center}
\tablestylesmaller{3.2pt}{1.05}
\begin{tabular}{lccccccccc}
\toprule
&   \multicolumn{3}{c}{CUB} & \multicolumn{3}{c}{Stanford Cars} & \multicolumn{3}{c}{FGVC-Aircraft}\\
\cmidrule(rl){2-4}\cmidrule(rl){5-7}\cmidrule(rl){8-10}
Methods        & All  & Old  & New  & All  & Old  & New  & All  & Old  & New \\
\midrule
$k$-means~\cite{macqueen1967some_kmeans}  & 34.3 & 38.9 & 32.1 & 12.8 & 10.6 & 13.8 & 16.0 & 14.4 & 16.8 \\
RS+~\cite{han21autonovel}          & 33.3 & 51.6 & 24.2 & 28.3 & 61.8 & 12.1 & 26.9 & 36.4 & 22.2 \\
UNO+~\cite{fini2021unified}               & 35.1 & 49.0 & 28.1 & 35.5 & 70.5 & 18.6 & 40.3 & 56.4 & 32.2 \\
ORCA~\cite{cao21orca}                     & 35.3 & 45.6 & 30.2 & 23.5 & 50.1 & 10.7 & 22.0 & 31.8 & 17.1 \\
\midrule
GCD~\cite{vaze22generalized}       & {51.3} & {56.6} & {48.7} & {39.0} & 57.6 & {29.9} & {45.0} & 41.1 & {46.9} \\
\OURS                     & \textbf{60.3} & \textbf{65.6} & \textbf{57.7} & \textbf{53.8} & \textbf{71.9} & \textbf{45.0} & \textbf{54.2} & \textbf{59.1} & \textbf{51.8} \\
$\Delta$                    & \textcolor{darkgreen}{\textbf{+9.0}} & \textcolor{darkgreen}{\textbf{+9.0}} & \textcolor{darkgreen}{\textbf{+9.0}} & \textcolor{darkgreen}{\textbf{+14.8}} & \textcolor{darkgreen}{\textbf{+14.3}} & \textcolor{darkgreen}{\textbf{+15.1}} & \textcolor{darkgreen}{\textbf{+9.2}} & \textcolor{darkgreen}{\textbf{+18.0}} & \textcolor{darkgreen}{\textbf{+4.9}} \\
\bottomrule
\end{tabular}
\end{center}
\vspace{-.8em}
\caption{Results on the Semantic Shift Benchmark~\cite{vaze22openset}.} \label{subtab:ssb}
\end{table}

\begin{table}[b]
\begin{center}
\tablestylesmaller{3.4pt}{1.05}
\begin{tabular}{lccccccccc}
\toprule
&    \multicolumn{3}{c}{CIFAR10} & \multicolumn{3}{c}{CIFAR100} & \multicolumn{3}{c}{ImageNet-100} \\
\cmidrule(rl){2-4}\cmidrule(rl){5-7}\cmidrule(rl){8-10}
Methods                                   & All  & Old  & New  & All  & Old  & New  & All  & Old  & New \\
\midrule
$k$-means~\cite{macqueen1967some_kmeans}  & 83.6 & 85.7 & 82.5 & 52.0 & 52.2 & 50.8 & 72.7 & 75.5 & {71.3} \\
RS+~\cite{han21autonovel}          & 46.8 & 19.2 & 60.5 & 58.2 & {77.6} & 19.3 & 37.1 & 61.6 & 24.8 \\
UNO+~\cite{fini2021unified}               & 68.6 & \textbf{98.3} & 53.8 & 69.5 & 80.6 & 47.2 & 70.3 & \textbf{95.0} & 57.9 \\
ORCA~\cite{cao21orca}                     & 81.8 & 86.2 & 79.6 & 69.0 & 77.4 & 52.0 & 73.5 & {92.6} & 63.9 \\
\midrule
GCD~\cite{vaze22generalized}       & {91.5} & {97.9} & {88.2} & {73.0} & 76.2 & {66.5} & {74.1} & 89.8 & 66.3 \\
\OURS                     & \textbf{97.1} & {95.1} & \textbf{98.1} & \textbf{80.1} & \textbf{81.2} & \textbf{77.8} & \textbf{83.0} & {93.1} & \textbf{77.9} \\
$\Delta$                  & \textcolor{darkgreen}{\textbf{+5.6}} & \textcolor{BrickRed}{\textbf{-2.8}} & \textcolor{darkgreen}{\textbf{+9.9}} & \textcolor{darkgreen}{\textbf{+7.1}} & \textcolor{darkgreen}{\textbf{+5.0}} & \textcolor{darkgreen}{\textbf{+11.3}} & \textcolor{darkgreen}{\textbf{+8.9}} & \textcolor{darkgreen}{\textbf{+3.3}} & \textcolor{darkgreen}{\textbf{+11.6}} \\
\bottomrule
\end{tabular}
\end{center}
\vspace{-.8em}
\caption{Results on generic image recognition datasets.} \label{subtab:generic}
\end{table}

\begin{table}[b]
\begin{center}
\tablestyle{7.5pt}{1}
\begin{tabular}{lcccccc}
\toprule
&    \multicolumn{3}{c}{Herbarium 19} & \multicolumn{3}{c}{ImageNet-1K}\\
\cmidrule(r){2-4}
\cmidrule(r){5-7}
Methods                                   & All  & Old  & New  & All  & Old  & New \\
\midrule
$k$-means~\cite{macqueen1967some_kmeans}  & 13.0 & 12.2 & 13.4 & - & - & - \\
RS+~\cite{han21autonovel}          & 27.9 & {55.8} & 12.8 & - & - & - \\
UNO+~\cite{fini2021unified}               & 28.3 & {53.7} & 14.7 & - & - & - \\
ORCA~\cite{cao21orca}                     & 20.9 & 30.9 & 15.5 & - & - & - \\
\midrule
GCD~\cite{vaze22generalized}       & {35.4} & 51.0 & {27.0} & 52.5 & 72.5 & 42.2 \\
\OURS                      & \textbf{44.0} & \textbf{58.0} & \textbf{36.4} & \textbf{57.1} & \textbf{77.3} & \textbf{46.9} \\ 
$\Delta$                  & \textcolor{darkgreen}{\textbf{+8.6}} & \textcolor{darkgreen}{\textbf{+7.0}} & \textcolor{darkgreen}{\textbf{+9.4}} & \textcolor{darkgreen}{\textbf{+4.6}} & \textcolor{darkgreen}{\textbf{+4.8}} & \textcolor{darkgreen}{\textbf{+4.7}} \\
\bottomrule
\end{tabular}
\end{center}
\vspace{-.8em}
\caption{Results on more challenging datasets.}\label{tab:herb19}
\end{table}

\begin{figure*}[t]
\centering
\includegraphics[width=\linewidth]{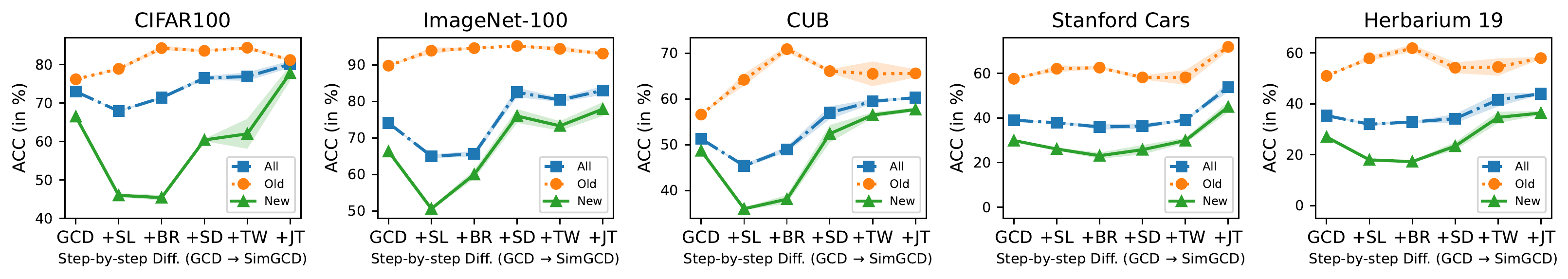}
\vspace{-1.6em}
\caption{
    \textbf{Step-by-step differences from GCD~\cite{vaze22generalized} to SimGCD.}
    (SL: self-labelling, BR: post-backbone representation, SD: self-distillation, TW: teacher temperature warmup, JT: joint training)
    }
\label{fig:ablation}
\end{figure*}

We compare with state-of-the-art methods in generalized category discovery (ORCA~\cite{cao21orca} and GCD~\cite{vaze22generalized}), strong baselines derived from novel category discovery (RS+~\cite{han21autonovel} and UNO+~\cite{fini2021unified}), and $k$-means~\cite{macqueen1967some_kmeans} on DINO~\cite{caron2021emerging} features.
On both the fine-grained SSB benchmark (\cref{subtab:ssb}) and generic image recognition datasets (\cref{subtab:generic}), our method achieves notable improvements in recognising `New' classes (across the instances in $\mathcal{D}^u$ that belong to classes in $\mathcal{Y}_u \text{\textbackslash} \mathcal{Y}_l$), outperforming the SOTAs by around 10\%.
The results in old classes are also competing against the best-performing baselines.
Given that the ability to discover `New' classes is a more desirable ability, the results are quite encouraging. 

In \cref{tab:herb19}, we also report the results on Herbarium 19~\cite{tan2019herbarium}, a naturally long-tailed fine-grained dataset that is closer to the real-world application of generalized category discovery, and ImageNet-1K~\cite{deng2009imagenet}, a large-scale generic classification dataset.
Still, our method shows consistent improvements in all metrics.

\begin{table}[htbp]
\begin{center}
\tablestyle{8pt}{1}
\begin{tabular}{lccccc}
\toprule
Methods      & CF100 & CUB & Herb19 & IN-100 & IN-1K \\
\midrule
GCD~\cite{vaze22generalized} & 7.5m & 9m & 2.5h & 36m & 7.7h \\
SimGCD & 1m & 18s & 3.5m & 9.5m & 0.6h \\
\bottomrule
\end{tabular}
\end{center}
\vspace{-.8em}
\caption{Inference time over the unlabelled split.} \label{tab:time}
\vspace{-.5em}
\end{table}

In \cref{tab:time}, we compare the inference time with GCD~\cite{vaze22generalized}, one iconic non-parametric classification method.
Let the number of all samples and unlabelled samples be $N$ and $N_u$, the number of classes $K$, feature dimension $d$, and the number of $k$-means iterations to be $t$, the time complexity of GCD is $\mathcal{O}(N^2d+NKdt)$ (including $k$-means++ initialisation), while our method only requires a nearest-neighbour prototype search for each instance, with time complexity $\mathcal{O}(N_uKd)$.
All methods adopt GPU implementations.

\subsection{Ablation Study} \label{subsec:ablation}
In \cref{fig:ablation}, we ablate the key components that bring the baseline method step-by-step to a new SOTA.

\paragraph{Baseline.}
We start from GCD~\cite{vaze22generalized}, a non-parametric classification framework.
We keep its representation learning objectives unchanged, and first impose the UNO~\cite{fini2021unified}-style self-labelling classification objectives (+SL) to it, thus transforming it into a parametric classifier.
The classifier is built on the projector, and detached from representation learning.
Results on `Old' classes generally improve, while results on `New' classes see a sharp drop. This is expected due to UNO's strong bias toward `Old' classes (\cref{fig:errors}).

\paragraph{Improving the representations.}
As suggested in \cref{subsec:whichrep}, we build the classifier on the backbone (+BR). %
This further makes notable improvements in `Old' classes, while changes in `New' classes vary across datasets. This indicates that the pseudo labels' quality is insufficient to benefit from the post-backbone representations (\cref{fig:clusterpos}).

\paragraph{Improving the pseudo labels.}
We start by replacing the self-labelling strategy with our self-distillation paradigm.
As shown in column (+SD), we achieve consistent improvements across all datasets by a large margin (\eg, 26\% in CIFAR100, 13\% in CUB) in `New' classes.
We then further adopt a teacher temperature warmup strategy (+TW) to lower the confidence of the pseudo-labels at an earlier stage.
The intuition is that at the beginning, both the classifier and the representation are not well fitted to the target data, thus the pseudo-labels are not quite reliable.
This is shown to be helpful for fine-grained classification datasets, while for generic classification datasets, which are similar to the pre-training data (ImageNet), the unreliable pseudo label is not a problem, thus lowering the confidence does not show help. For simplicity, we keep the training strategy consistent.

\paragraph{Jointly training the representation.}
Previous settings adopt a decoupled training strategy for consistent representations with GCD~\cite{vaze22generalized} and fair comparison.
Finally, as confirmed in \cref{subsec:decouple}, we jointly supervise the representation with the classification objective (+JT).
This results in a consistent improvement in `New' classes for all datasets.
Changes in `Old' classes are mostly neutral or positive, with a notable drop in CIFAR100. Our intuition is that the original representations are already good enough for `Old' classes in this dataset, and some incorrect pseudo labels lead to sight degradation in this case.

\begin{figure}[b]
\centering
\includegraphics[width=\linewidth]{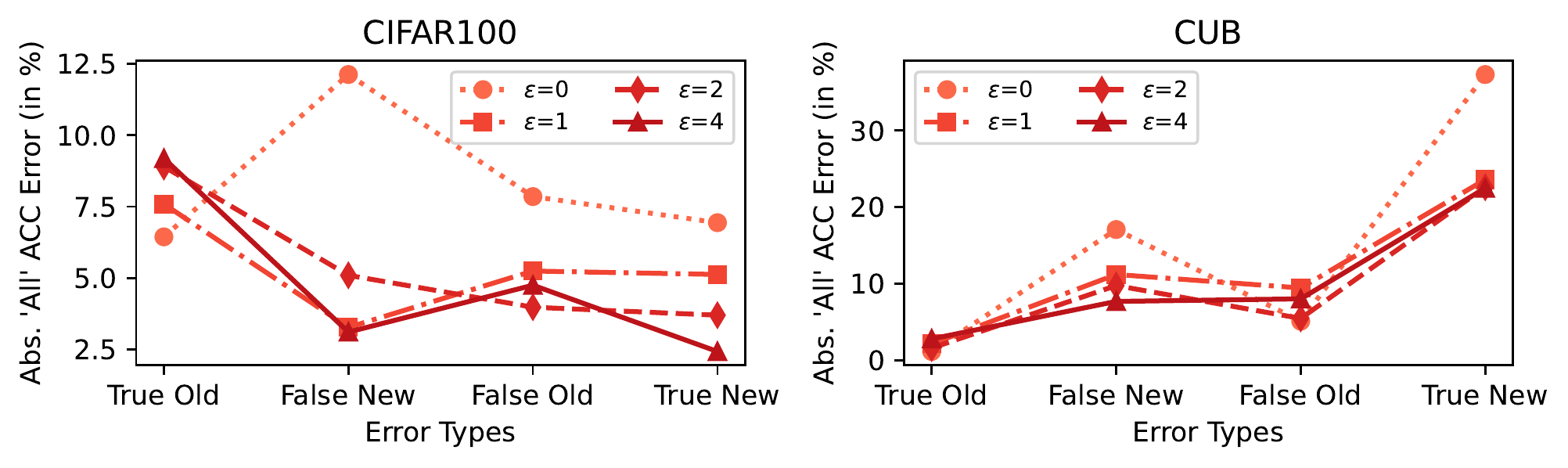}
\vspace{-1.6em}
\caption{
    \textbf {Effect of entropy regularisation on four types of classification errors.} 
    Appropriate entropy regularisation helps overcome the bias between `Old'/`New' classes (see ``False New'' and ``False Old'', lower is better).
    }
\label{fig:ours_error}
\end{figure}
\begin{figure}[b]
\centering
\includegraphics[width=\linewidth]{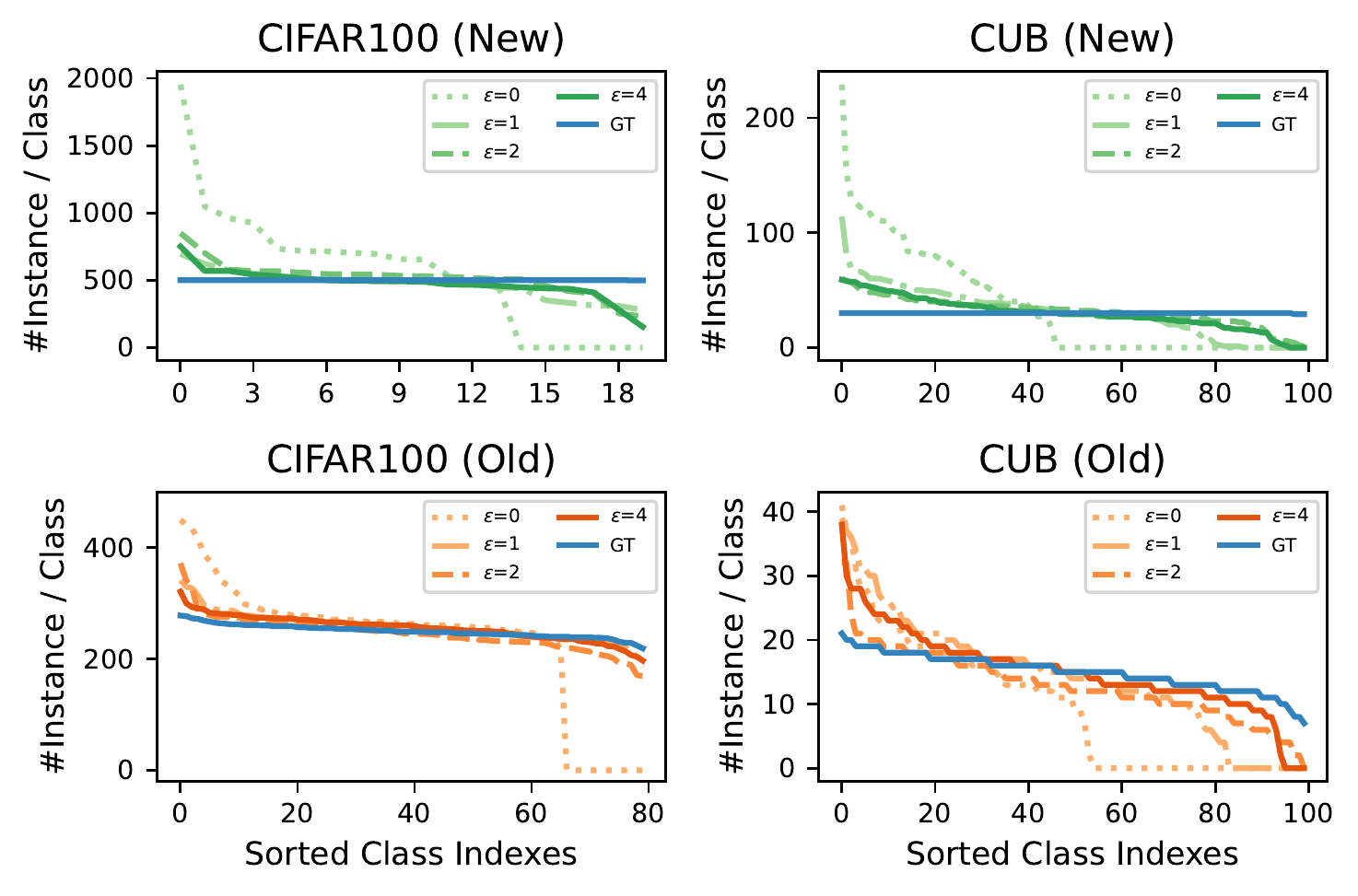}
\vspace{-1.6em}
\caption{
\textbf{Per-class prediction distributions with different entropy regularisation weights.} 
Proper entropy regularisation helps overcome the bias across `Old'/`New' classes, and approach the GT class distribution.
}
\label{fig:ours_longtail}
\end{figure}

\begin{figure*}[t]
\centering
\includegraphics[width=\linewidth]{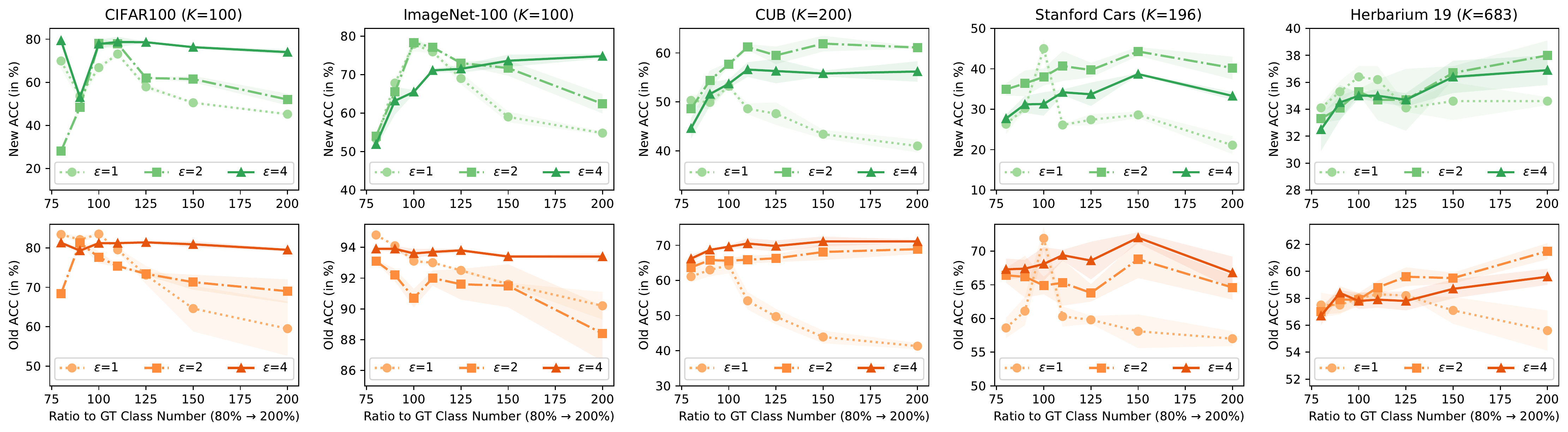}
\caption{
    \textbf{Results with different numbers of categories.}
    Stronger entropy regularisation effectively enforces the model's robustness to unknown numbers of categories, but over-regularisation may limit the ability to recognise `New' classes under ground-truth class numbers.
    }
\label{fig:ks}
\end{figure*}

\subsection{Analyses And Discussions}\label{subsec:discuss}

\paragraph{Entropy regularisation helps overcome prediction bias.}
We verify the effectiveness of entropy regularisation in overcoming prediction bias by diagnosing the model's classification errors and class-wise prediction distributions.
\cref{fig:ours_error} shows that this term consistently helps reduce ``False New'' and ``False Old'' errors, which refer to predicting an `Old' class sample to a `New' class, and vice-versa.
Besides, \cref{fig:ours_longtail} shows proper entropy regularisation helps overcome the imbalanced pseudo labels across all classes, and approach the ground truth (GT) class distribution.

\begin{figure}[b]
\centering
\vspace{-.5em}
\includegraphics[width=\linewidth]{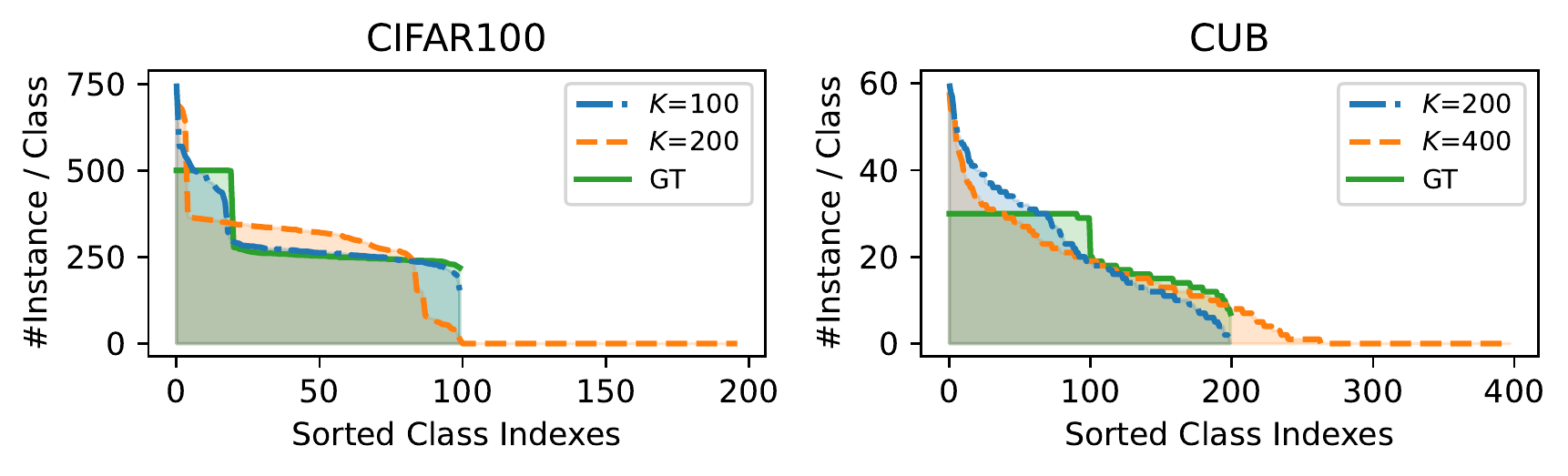}
\caption{
\textbf{Per-class prediction distributions with different numbers of categories.} 
Our method effectively identifies the criterion for `New' classes, thus keeping the number of active prototypes close to the ground-truth class number.
} \label{fig:largeK_analysis}
\end{figure}

\paragraph{Entropy regularisation enforces robustness to unknown numbers of categories.}
The main text assumed the category number is known \textit{a-priori} following prior works~\cite{han21autonovel,zhao21novel,zhong2021openmix,fini2021unified}, which is impractical~\cite{zhao2023gpc}.
In \cref{fig:ks}, we present the results with different numbers of categories on five representative datasets.
A category number lower than the ground truth significantly limits the ability to discover `New' categories, and the model tends to focus more on the `Old' classes.
On the other hand, increasing the category number results in less harm to the generic image recognition datasets and can even be helpful for some datasets.
When a stronger entropy penalty is imposed, the model shows strong robustness to the category number. Interestingly, further analysis in \cref{fig:largeK_analysis} shows the network prefers to keep the number of active prototypes low and close to the real category number. This finding is inspiring and could ease the deployment of GCD in real-world scenarios.

\begin{figure}[htbp]
\centering
\includegraphics[width=\linewidth]{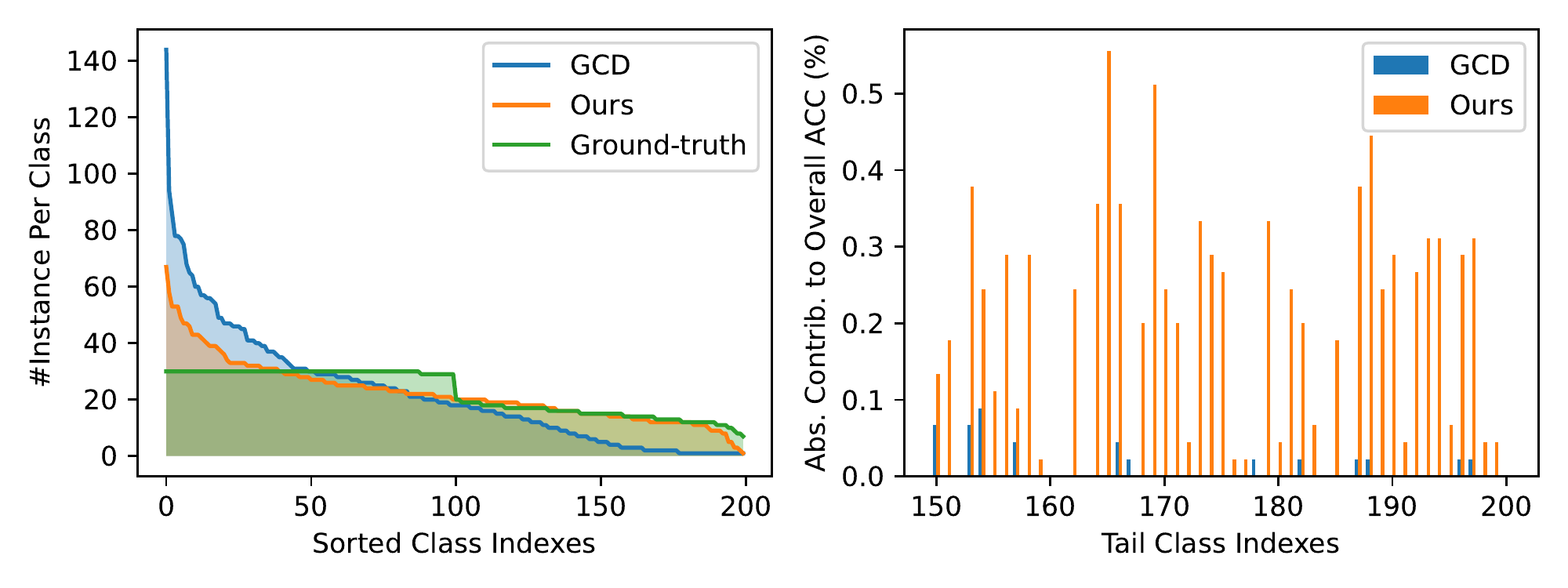}
\caption{
    \textbf{Prediction analysis against GCD~\cite{vaze22generalized}.}
    Left: Based on identical representations, the non-parametric classifier (semi-supervised $k$-means) adopted by GCD produces highly imbalanced predictions, while our method better fits the true distribution;
    Right: our method significantly improves GCD's tail classes.
}
\label{fig:error}
\end{figure}

\paragraph{What makes for the significant improvements over GCD given identical representations?}
One interesting message from \cref{fig:ablation} is that, even with the same representations (col. +TW), we can already improve GCD by a large margin.
We thus study the classification predictions and the major components that lead to the performance gap.
As shown in \cref{fig:error}, the non-parametric classifier (semi-supervised $k$-means) adopted by GCD~\cite{vaze22generalized} produces highly imbalanced predictions, while our method better fits the true distribution.
Further analysis (right part) shows that our method significantly improves over the tail classes of GCD.

\begin{figure}[htbp]
\centering
\includegraphics[width=\linewidth]{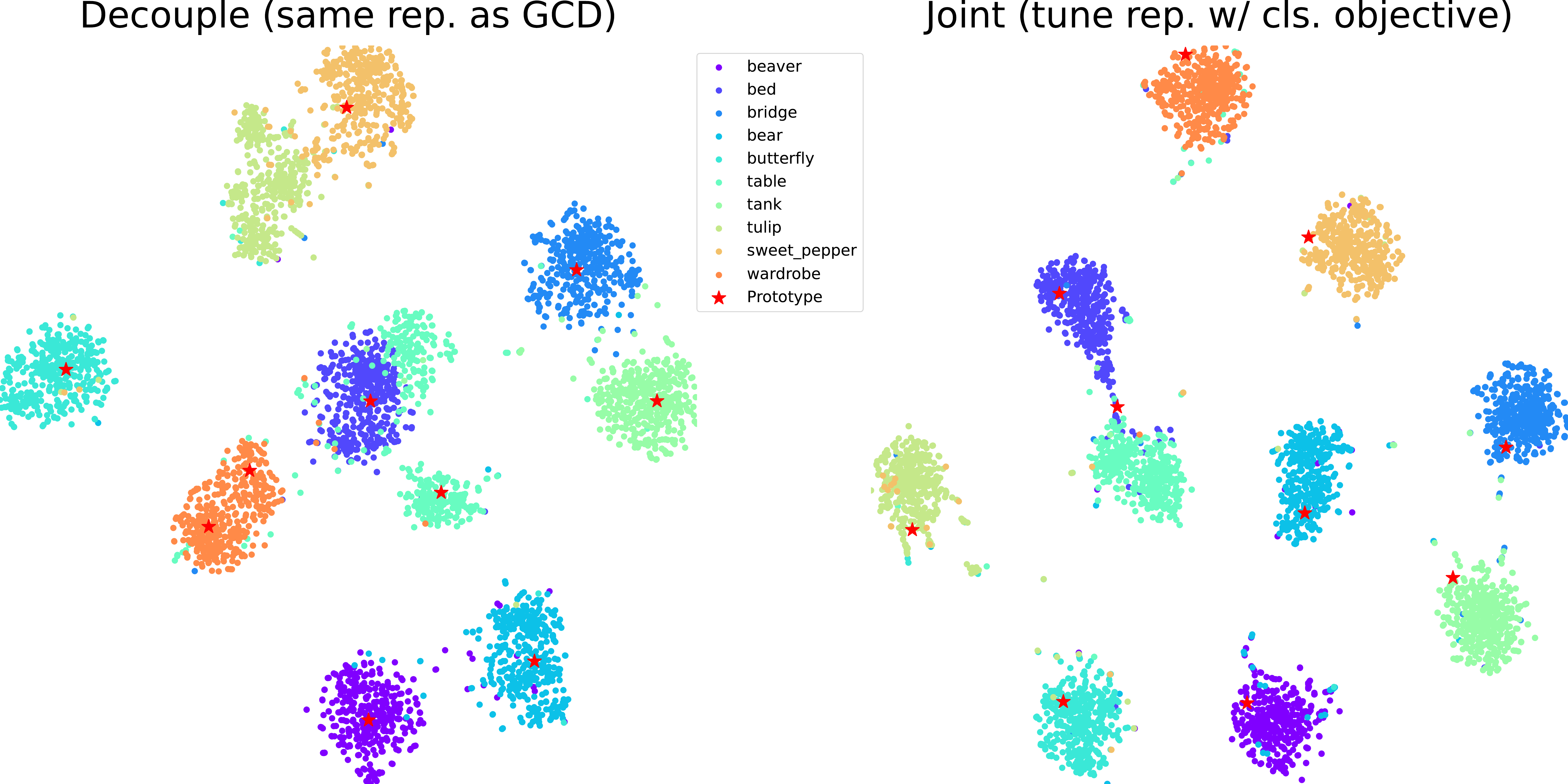}
\caption{
    \textbf{T-SNE~\cite{van2008visualizing_tsne} visualisation of the representations of 10 classes randomly sampled from CIFAR100~\cite{cifar}.}
    Jointly supervising representation learning with a classification objective helps disambiguate (\eg, bed \& table) and forms compacter clusters. 
    }
\label{fig:tsne}
\end{figure}

\paragraph{How does the classification objective change the representations?}
In \cref{fig:ablation}, we have shown that jointly training the representations with the classification objective can lead to $\sim$15\% boost in `New' classes on CIFAR100.
We study this difference by visualising the representations before and after tuning with t-SNE~\cite{van2008visualizing_tsne}.
As in \cref{fig:tsne}, jointly tuning the feature leads to less ambiguity, larger margins, and compacter clusters.
Concerning why this is not as helpful for CUB: we hypothesise that one important factor lies in how transferable the features learned in `Old' classes are to `New' classes. While it may be easier for a cat classifier to be adapted to dogs, things can be different for fine-grained bird recognition.
Besides, the small scale of CUB, which contains only 6k images while holding a large class split (200), might also make it hard to learn transferable features.

\begin{figure}[htbp]
\centering
\includegraphics[width=\linewidth]{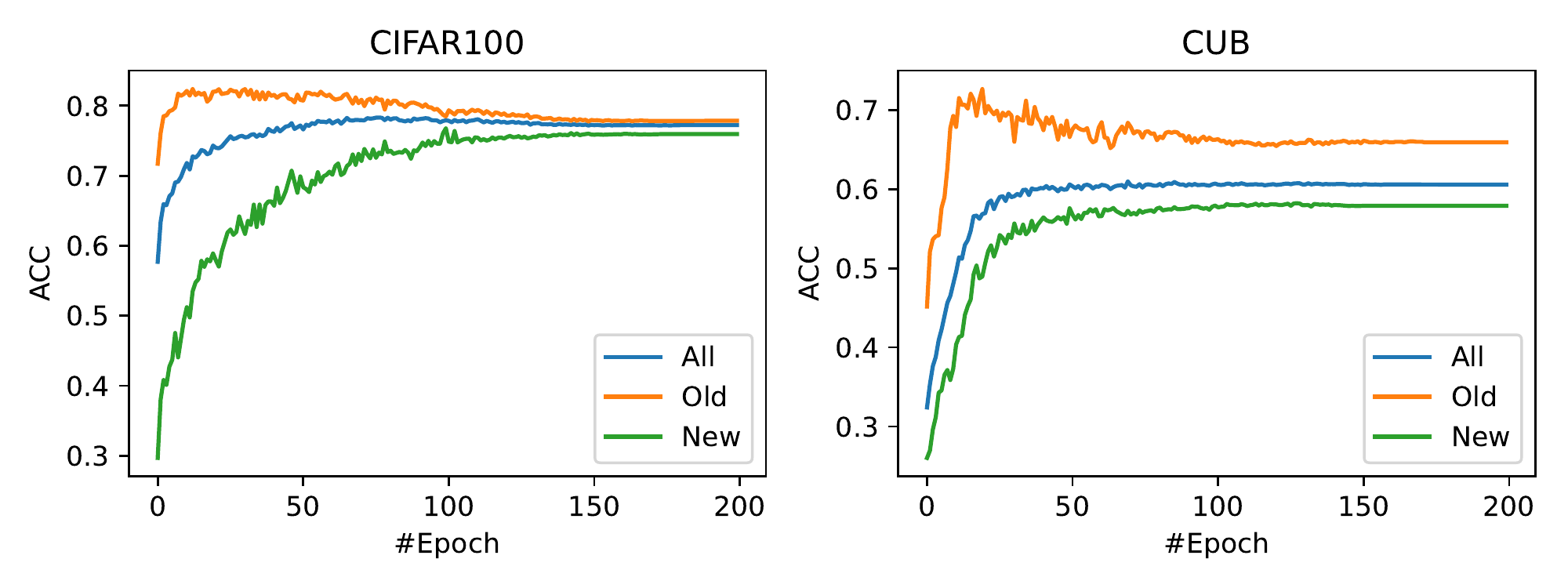}
\caption{
    \textbf{Performance evolution throughout the model learning process.}
    We observe a trade-off between the performance in `Old' and `New' categories, which is common across datasets.
    }
\label{fig:tradeoff}
\end{figure}
\paragraph{Trade-off between `Old' and `New' categories.}
We plot the performance evolution throughout the model learning process in~\cref{fig:tradeoff}. It can be observed that the performance on the `Old' categories first climbs to the highest point at the early stage of training and then slowly degrades as the performance on the `New' categories improves. 
We believe this demonstrates an important aspect of the design of models for the GCD problem: the performance on the `Old' categories may be in odd with the performance on the `New' categories, how to achieve a better trade-off between these two could be an interesting investigation for future works.

\section{Limitations and Potential Future Works}

\paragraph{Representation learning.}
This paper mainly targets improving the classification ability for generalized category discovery. The representation learning, however, follows the prior work GCD~\cite{vaze22generalized}.
It is expectable that the quality of representation learning can be improved. For instance, generally, by using more advanced geometric and photometric data augmentations~\cite{grill2020bootstrap}, and even multiple local crops~\cite{caron2020unsupervised}. Further, can the design of data augmentations be better aligned with the classification criteria of the target data?
For another example, using a large batch size has been shown to be critical to the performance of contrastive learning-based frameworks~\cite{chen2020simple}. However, the batch size adopted by GCD~\cite{vaze22generalized} is only 128, which might limit the quality of learned representations.
Moreover, is the supervised contrastive learning plus self-supervised contrastive learning paradigm the ultimate answer to form the feature manifold? We believe that advances in representation learning can lead to further gains.

\paragraph{Alignment to human-defined categories.}
This paper follows the common practice of previous works where human labels in seen categories implicitly define the metric for unseen ones, which can be viewed as an effort to align algorithm-discovered categories with human-defined ones.
However, labels in seen categories may not be good guidance when there is a gap between seen ones and the novel categories we want to discover, \eg, how to use the labelled images in ImageNet to discover novel categories in CUB? For another example, when we use a very big class vocabulary (\eg, the full ImageNet-22K~\cite{deng2009imagenet}), categories could overlap with each other, and be in different granularities. Further, assigning text names to the discovered categories still requires a matching process, what if further utilising the relationship between class names, and directly predicting the novel categories in the text space?
We believe the alignment between algorithm-discovered categories and human-defined categories is of high research value for future works.

\paragraph{Ethical considerations.}
Current methods commonly suffer from low-data or long-tailed scenarios. Depending on the data and classification criteria of specific tasks, discrimination against minority categories or instances is possible.
\section{Conclusion}
This study investigates the reasons behind the failure of previous parametric classifiers in recognizing novel classes in GCD and uncovers that unreliable pseudo-labels, which exhibit significant biases, are the crucial factor.
We propose a simple yet effective parametric classification method that addresses these issues and achieves state-of-the-art performance on multiple GCD benchmarks.
Our findings provide insights into the design of robust classifiers for discovering novel categories and we hope our proposed framework will serve as a strong baseline to facilitate future studies in this field and contribute to the development of more accurate and reliable methods for category discovery.

\section*{Acknowledgements}
This work has been supported by Hong Kong Research Grant Council - Early Career Scheme (Grant No. 27209621), General Research Fund Scheme (Grant No. 17202422), and RGC Matching Fund Scheme (RMGS). Part of the described research work is conducted in the JC STEM Lab of Robotics for Soft Materials funded by The Hong Kong Jockey Club Charities Trust.
The authors acknowledge SmartMore and MEGVII for partial computing support, and Zhisheng Zhong for professional suggestions.

{\small
\bibliographystyle{ieee_fullname}
\bibliography{simgcd}
}
\clearpage

\appendix

\twocolumn[
\centering
\vspace{3.7em}
{\Large
\textbf{Parametric Classification for Generalized Category Discovery: A Baseline Study}\\
\vspace{.5em}
Supplementary Material
}\\
\vspace{2.4em}
\author{%
  \large
  Xin Wen\textsuperscript{$1$*} \qquad
  Bingchen Zhao\textsuperscript{$2$*} \qquad
  Xiaojuan Qi\textsuperscript{$1$} \\
  \textsuperscript{$1$}The University of Hong Kong \quad
  \textsuperscript{$2$}University of Edinburgh \\
  \tt\small \{wenxin,xjqi\}@eee.hku.hk \qquad
  \tt bingchen.zhao@ed.ac.uk \\
}
\vspace{3em}
] %

{
  \hypersetup{linkcolor=black}
  \tableofcontents
}

\addtocontents{toc}{\protect\setcounter{tocdepth}{2}}

\section{Implementation Details}
\subsection{Experiment Setting Details}
The split of labelled (`Old') and unlabelled (`New') categories follows GCD~\cite{vaze22generalized}. That is, 50\% of all classes are sampled as `Old' classes ($\mathcal{Y}_l$), and the rest are regarded as `New' classes ($\mathcal{Y}_u \setminus \mathcal{Y}_l$). The exception is CIFAR100, for which 80\% classes are sampled as `Old',
following the novel category discovery (NCD) literature.
Regarding the sampling process, for generic object recognition datasets, the labelled classes are selected by their class index (the first $|\mathcal{Y}_l|$ ones). For the Semantic Shift Benchmark, data splits provided in~\cite{vaze22openset} are adopted. For Herbarium 19~\cite{tan2019herbarium}, the labelled classes are sampled randomly.
Additionally, for ImageNet-1K~\cite{deng2009imagenet} which is not used in~\cite{vaze22generalized}, we follow its fashion to select the first 500 classes sorted by class id as the labelled classes.
Then for all datasets, following~\cite{vaze22generalized}, 50\% of the images from the labelled classes are randomly sampled to form the labelled dataset $\mathcal{D}^l$, and all remaining images are regarded as the unlabelled dataset $\mathcal{D}^u$.
All experiments are done with a batch size of 128 on a single GPU, except for ImageNet-1K, on which we train with eight GPUs, scale the learning rate with the linear scaling rule, and keep the per-GPU batch size unchanged. The inference time on ImageNet-1K is still evaluated with one GPU.

\subsection{Re-implementing Previous Works}

Results of GCD~\cite{vaze22generalized} are taken from the original paper (if available), and otherwise re-implemented with the official codebase.
One exception is ImageNet-1K~\cite{deng2009imagenet}, which was not evaluated by the authors. %
Since naively adopting their official codebase to ImageNet-1K fails as the semi-supervised $k$-means procedure requires too much GPU memory and cannot be done with available hardware, we drop the $k$-mean++ initialisation~\cite{Arthur2008kmeanspp} which takes the most memory, and re-implement the method with faiss~\cite{johnson2019billion} for speed up (otherwise the evaluation takes more than one day).
The results are in the main paper, compared to our proposed strong baseline SimGCD, GCD requires significantly more time to run and more engineering efforts, and yet achieves a lower performance than SimGCD, which demonstrates the effectiveness of our proposed method.
Results of UNO+~\cite{fini2021unified} and RS+~\cite{rankstat}, which are adaptations of the original works to the GCD task, are directly taken from the GCD~\cite{vaze22generalized} paper.
Also note that unlike UNO~\cite{fini2021unified}, our method does not adopt the over-clustering trick for simplicity.
Results of ORCA~\cite{cao21orca} are re-implemented with the official codebase. We align the details in dataset split and backbone (ViT-B/16~\cite{dosovitskiy2020vit} pre-trained with DINO~\cite{caron2021emerging}) with GCD~\cite{vaze22generalized} for a fair comparison.

\begin{figure*}[t]
\centering
\includegraphics[width=\linewidth]{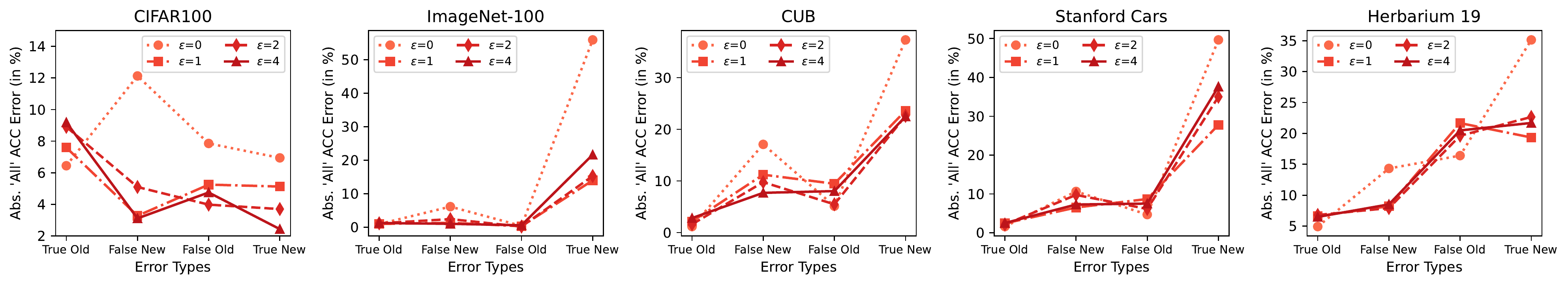}
\vspace{-1.5em}
\caption{
\textbf{Complete error analysis results of SimGCD on five representative datasets.} 
With appropriate entropy regularisation, the bias between `Old'/`New' classes (see ``False New'' and ``False Old'' errors) are generally effectively alleviated, except in the long-tailed Herbarium 19 that the effect varies. Also notably, ``True New'' errors are consistently penalised to a considerable extent, confirming entropy regularisation's ability in helping recognise and distinguish between novel categories.
} \label{fig:errors_ours_full}
\end{figure*}

\subsection{Error Analysis Details}

We briefly clarify the details of obtaining the four kinds of prediction errors in the main paper: we first rank the category indexes in consecutive order, such that by index, all `Old' classes are followed by all `New' classes. We then compute the full confusion matrix, with each element summarising how many times images of one specific class (row index) are predicted as one class (column index). All elements are divided by the number of testing samples to account for the percentage. We then reduce the diagonal terms to zero (representing correct predictions), and thus all remaining elements represent different kinds of prediction errors (\ie, absolute contribution to the errors of `All' ACC). Finally, we slice the confusion matrix into four sub-matrices at the boundaries between the `Old' and `New' classes, and add all elements in each sub-matrix together, thus obtaining the final error matrix standing for the four kinds of prediction errors.
Such a way of error classification helps distinguish the prediction bias between and within seen and novel categories, and thus facilitates the design of new solutions.
Note that the diagonal elements, \eg, `True Old' predictions, do not stand for correct predictions, but for cases that incorrectly predicting samples of one specific `Old' class to another wrong `Old' class.

\section{Extended Experiments And Discussions}
\subsection{Main Results}

We present the full results of SimGCD in the main paper with error bars in \cref{tab:main_complete}. The results are obtained from three independent runs and thus avoid randomness.

\begin{table}[htbp]
\centering
\tablestyle{9pt}{1.05}
\begin{tabular}{lccc}
    \toprule
    Dataset     & All   & Old   & New \\
    \midrule
    CIFAR10~\cite{cifar}     & 97.1$\pm$0.0     & 95.1$\pm$0.1     & 98.1$\pm$0.1  \\
    CIFAR100~\cite{cifar}    & 80.1$\pm$0.9     & 81.2$\pm$0.4     & 77.8$\pm$2.0  \\
    ImageNet-100~\cite{deng2009imagenet}  & 83.0$\pm$1.2     & 93.1$\pm$0.2     & 77.9$\pm$1.9  \\
    ImageNet-1K~\cite{deng2009imagenet}   & 57.1$\pm$0.1     & 77.3$\pm$0.1     & 46.9$\pm$0.2  \\
    CUB~\cite{cub200}        & 60.3$\pm$0.1     & 65.6$\pm$0.9     & 57.7$\pm$0.4  \\
    Stanford Cars~\cite{stanfordcars}     & 53.8$\pm$2.2     & 71.9$\pm$1.7     & 45.0$\pm$2.4  \\  
    FGVC-Aircraft~\cite{aircraft}         & 54.2$\pm$1.9     & 59.1$\pm$1.2     & 51.8$\pm$2.3  \\
    Herbarium 19~\cite{tan2019herbarium}  & 44.0$\pm$0.4     & 58.0$\pm$0.4     & 36.4$\pm$0.8  \\
    \bottomrule
\end{tabular}
\vspace{.5em}
\caption{Complete results of SimGCD in three independent runs.}\label{tab:main_complete}
\vspace{-.2em}
\end{table}

\subsection{Unknown Category Number}
In the main text, we showed that the performance of SimGCD is robust to a wide range of estimated unknown category numbers. 
In this section, we report the results with the number of categories estimated using an off-the-shelf method~\cite{vaze22generalized} (\cref{tab:predicted_k}) or with a roughly estimated relatively big number (two times of the ground-truth $K$), and compare with the baseline method GCD~\cite{vaze22generalized}.

\begin{table}[ht]
    \centering
    \tablestyle{5pt}{1.05}
    \begin{tabular}{lccccc}
    \toprule
       & CIFAR100   & ImageNet-100   & CUB    & SCars & Herb19 \\
    \midrule
    GT   $K$      & 100 & 100 & 200 & 196 & 683\\
    Est. $K$      & 100 & 109 & 231 & 230 & 520\\
    \bottomrule
    \end{tabular}
    \vspace{.5em}
    \caption{Number of categories $K$ estimated using~\cite{vaze22generalized}.}\label{tab:predicted_k}
\end{table}

\begin{table}[ht]
\vspace{.5em}
\centering
\tablestyle{4.5pt}{1.05}
\begin{tabular}{lcccccccccc}
\toprule
&   &   \multicolumn{3}{c}{CIFAR100} & \multicolumn{3}{c}{ImageNet-100}\\
\cmidrule(rl){3-5}
\cmidrule(rl){6-8}
Methods          & \makecell{Known $K$} & All  & Old  & New  & All  & Old  & New \\\midrule
GCD~\cite{vaze22generalized}   & \cmark & 73.0 & 76.2 & 66.5 & {74.1} & {89.8} & {66.3} \\
\OURS                     & \cmark & \textbf{80.1} & \textbf{81.2} & \textbf{77.8} & \textbf{83.0} & \textbf{93.1} & \textbf{{77.9}} \\
\midrule
\rowcolor{baselinecolor}GCD~\cite{vaze22generalized}          & \xmark\,(w/ Est.) & 73.0 & 76.2 & 66.5 & 72.7 & 91.8 & 63.8 \\
\rowcolor{baselinecolor}\OURS                        & \xmark\,(w/ Est.) & \textbf{80.1} & \textbf{81.2} & \textbf{77.8} & \textbf{81.7} & 91.2 & \textbf{76.8} \\
\rowcolor{baselinecolor}\OURS        & \xmark\,(w/ $2K$)  & 77.7 & 79.5 & 74.0 & 80.9 & \textbf{93.4} & 74.8 \\
\bottomrule
\end{tabular}
\vspace{.5em}
\caption{Results on generic image recognition datasets.}\label{tab:unknownk_generic}
\end{table}

\begin{table}[ht]
\centering
\tablestyle{4.5pt}{1.05}
\begin{tabular}{lcccccccccc}
\toprule
&   &   \multicolumn{3}{c}{CUB} & \multicolumn{3}{c}{Stanford Cars}\\
\cmidrule(rl){3-5}
\cmidrule(rl){6-8}
Methods                                   & \makecell{Known $K$} & All  & Old  & New  & All  & Old  & New \\\midrule
GCD~\cite{vaze22generalized}      & \cmark & 51.3 & 56.6 & 48.7 & {39.0} & {57.6} & {29.9} \\
\OURS                     & \cmark & \textbf{60.3} & \textbf{65.6} & \textbf{57.7} & \textbf{53.8} & \textbf{71.9} & \textbf{45.0} \\
\midrule
\rowcolor{baselinecolor}GCD~\cite{vaze22generalized}                 & \xmark\,(w/ Est.) & 47.1 & 55.1 & 44.8 & 35.0 & 56.0 & 24.8 \\
\rowcolor{baselinecolor}\OURS             & \xmark\,(w/ Est.) & 61.5 & 66.4 & 59.1 & \textbf{49.1} & \textbf{65.1} & \textbf{41.3} \\
\rowcolor{baselinecolor}\OURS        & \xmark\,(w/ $2K$)  & \textbf{63.6} & \textbf{68.9} & \textbf{61.1} & 48.2 & 64.6 & 40.2 \\
\bottomrule
\end{tabular}
\vspace{.5em}
\caption{Results on the Semantic Shift Benchmark~\cite{vaze22openset}.}\label{tab:unknownk_ssb}
\end{table}

The results on CIFAR100~\cite{cifar}, ImageNet-100~\cite{deng2009imagenet}, CUB~\cite{cub200}, and Stanford Cars~\cite{stanfordcars} are available in \cref{tab:unknownk_generic,tab:unknownk_ssb}.
Our method shows consistent improvements on four representative datasets when $K$ is unknown, no matter with the category number estimated with a specialised algorithm (w/ Est.), or simply with a loose estimation that is two times the ground truth (w/ $2K$, other values are also applicable since our method is robust to a wide range of estimations). This property could ease the deployment of parametric classifiers for GCD in real-world scenarios.

\begin{figure*}[htbp]
\centering
\includegraphics[width=\linewidth]{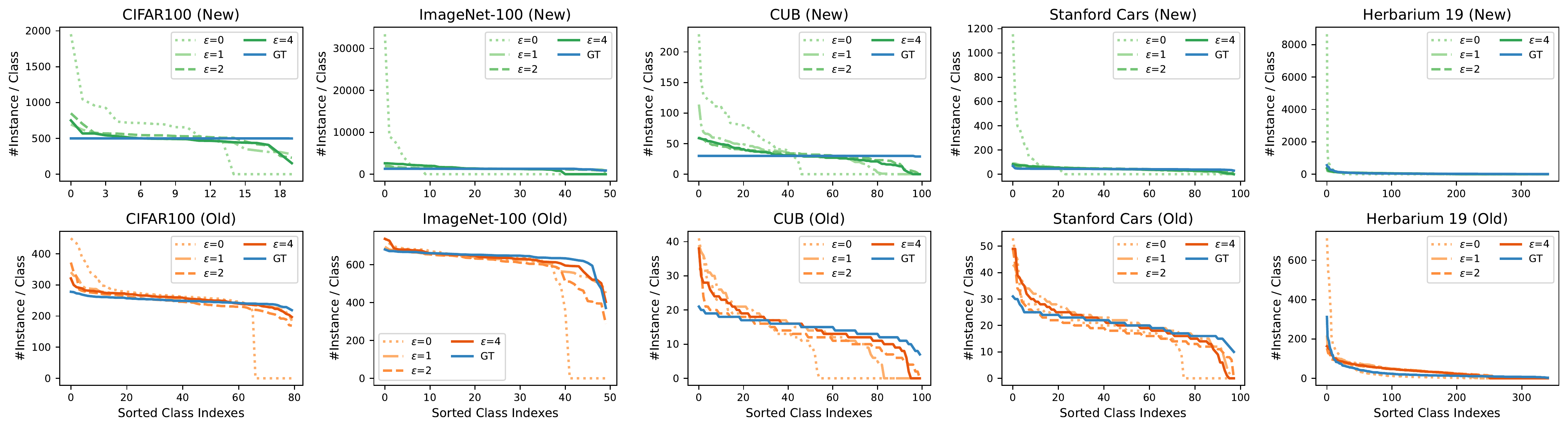}
\caption{
\textbf{Complete per-class prediction distribution results of SimGCD on five representative datasets.} 
Proper entropy regularisation helps overcome the prediction bias in both `Old' classes and `New' classes, and fits the ground-truth distribution. The conclusion is consistent across generic classification datasets, fine-grained classification datasets, and naturally long-tailed datasets. 
} \label{fig:ours_longtail_full}
\end{figure*}

\subsection{Extended Analyses}

In supplementary to the main paper, we present a more complete version of the analytical experiments.

In \cref{fig:errors_ours_full}, we show the error analysis results of SimGCD over five representative datasets that cover coarse-grained, fine-grained, and long-tailed classification tasks. Overall, it shows that the entropy regulariser mainly helps in overcoming two types of errors: the error of misclassification between `Old'/`New' categories, and the error of misclassification within `New' categories.
One exception is the long-tailed Herbarium 19 dataset, in which the models' ``False Old'' errors also increased, and our intuition is that the long-tailed distribution adds to the difficulty in discriminating between `Old' and `New' categories. Still, the gain in distinguishing between novel categories is consistent, and we provide a further analysis via per-class prediction distributions in the next paragraph.

\begin{figure}[htbp]
\centering
\includegraphics[width=\linewidth]{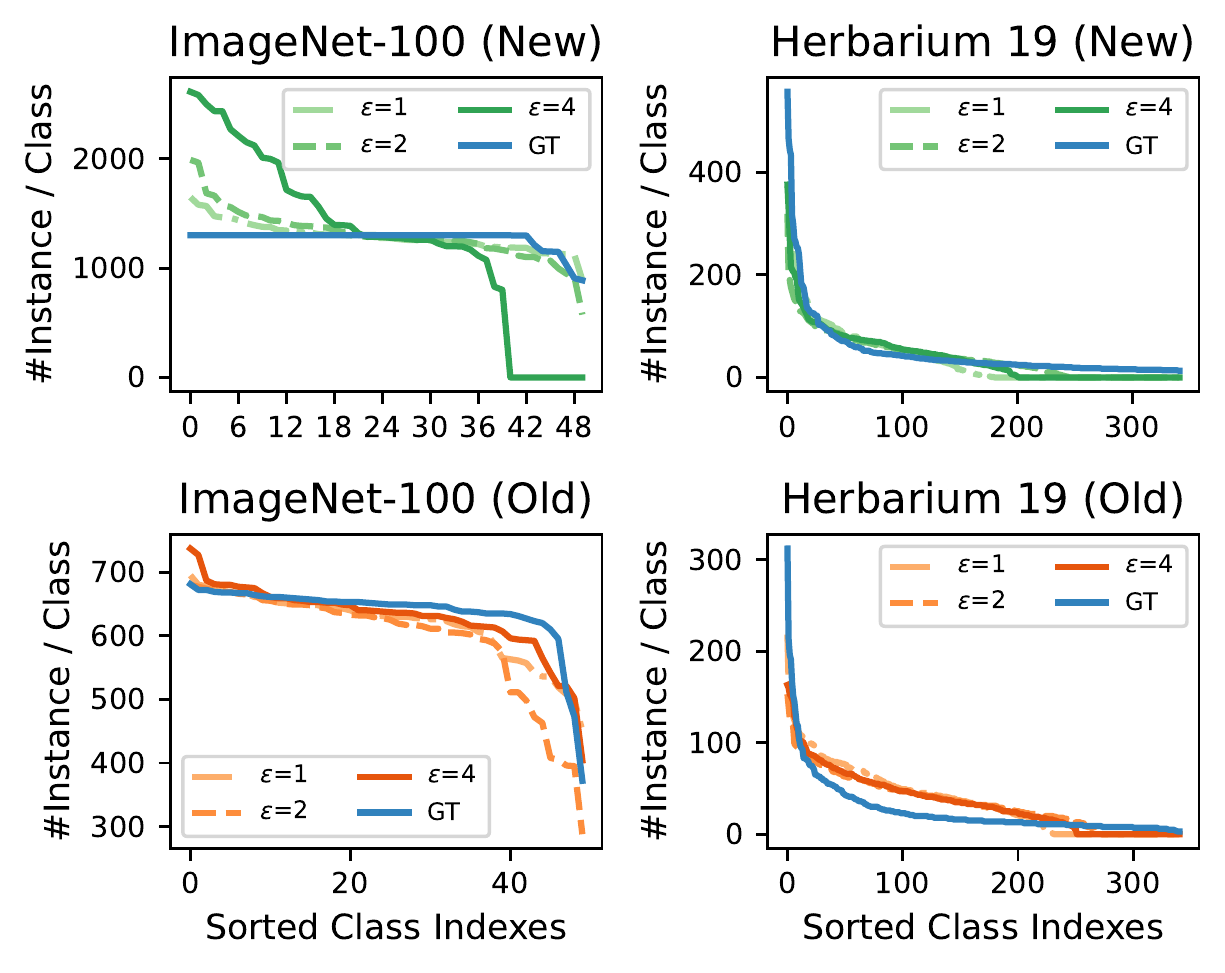}
\caption{
\textbf{A closer look at the per-class distributions.} 
Notably, although the entropy regularisation term is formulated to approach uniform distribution, it could make the models' predictions more biased on the class-balanced ImageNet-100 dataset when the regularisation is too strong. Interestingly, it also could help fit the distribution of the long-tailed Herbarium 19 dataset.
} \label{fig:ours_longtail_closer}
\end{figure}

In \cref{fig:ours_longtail_full}, we show the complete per-class prediction results of SimGCD to further analyse the entropy regulariser's effect in overcoming the classification errors within `Old' and `New' classes, and it consistently verifies the help in alleviating the prediction bias within `Old' and `New' classes, and better fitting the ground-truth class distribution.
In \cref{fig:ours_longtail_closer}, we present a closer look at ImageNet-100 and Herbarium 19.
The entropy regularisation term is formulated to make the model's predictions closer to the uniform distribution. But interestingly, we empirically found that it could make the models' predictions more biased on the class-balanced ImageNet-100 dataset when the regularisation is too strong. And when the dataset itself is long-tailed (Herbarium 19), it also could help fit the ground-truth distribution. We also note that the self-labelling strategy adopted by UNO~\cite{fini2021unified} forces the predictions in a batch to be strictly uniform, which may account for its inferior performance.

\begin{figure}[htbp]
\centering
\includegraphics[width=\linewidth]{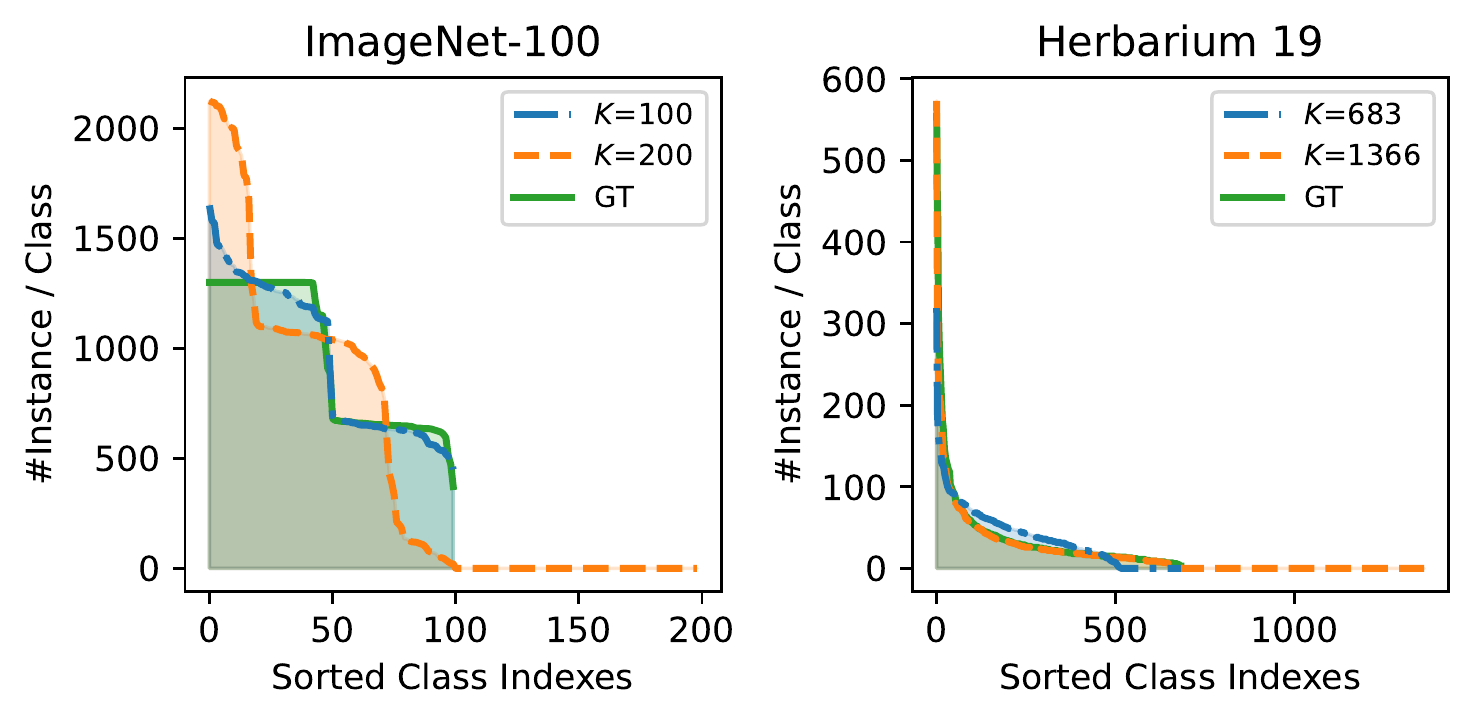}
\caption{
\textbf{Per-class prediction distributions using different category numbers on ImageNet-100 and Herbarium 19.} 
Our method effectively identifies the criterion for `New' classes, thus keeping the number of active prototypes close to the ground-truth class number.
Notably, a loose category number greater than the ground truth may harm fitting the class-balanced ImageNet-100 dataset, but could help fit the distribution of the long-tailed Herbarium 19 dataset.
} \label{fig:largeK_analysis_another}
\end{figure}

\begin{table*}[t]
\centering
\tablestyle{5.7pt}{1.1}
\begin{tabular}{lcccccccccccccccc}
\toprule
& & \multicolumn{3}{c}{CIFAR100} & \multicolumn{3}{c}{ImageNet-100} & \multicolumn{3}{c}{CUB} & \multicolumn{3}{c}{Stanford Cars} & \multicolumn{3}{c}{Herbarium 19} \\
\cmidrule(rl){3-5} \cmidrule(rl){6-8} \cmidrule(rl){9-11} \cmidrule(rl){12-14} \cmidrule(rl){15-17}
Method & Logit Adjust & All & Old & New & All & Old & New & All & Old & New & All & Old & New & All & Old & New \\
\midrule
ORCA~\cite{cao21orca}      & \cmark & 69.0 & 77.4 & 52.0 & 73.5 & 92.6 & 63.9 & 35.3 & 45.6 & 30.2 & 23.5 & 50.1 & 10.7 & 20.9 & 30.9 & 15.5 \\
DebiasPL~\cite{wang2022debiased}  & \cmark & 60.9 & 69.8 & 43.1 & 43.5 & 59.1 & 35.6 & 38.1 & 44.2 & 35.0 & 31.1 & 49.6 & 22.1 & 30.1 & 39.1 & 25.3 \\
\midrule
\rowcolor{baselinecolor}UNO+~\cite{fini2021unified}     & \xmark &69.5 & 80.6 & 47.2 & 70.3 & \textbf{95.0} & 57.9 & 35.1 & 49.0 & 28.1 & 35.5 & 70.5 & 18.6 & 28.3 & 53.7 & 14.7 \\
\rowcolor{baselinecolor}GCD~\cite{vaze22generalized}       & \xmark & 73.0 & 76.2 & 66.5 & 74.1 & 89.8 & 66.3 & 51.3 & 56.6 & 48.7 & 39.0 & 57.6 & 29.9 & 35.4 & 51.0 & 27.0 \\
\rowcolor{baselinecolor}SimGCD    & \xmark & \textbf{80.1} & \textbf{81.2} &  \textbf{77.8} & \textbf{83.0} & 93.1 &  \textbf{77.9} & \textbf{60.3} & \textbf{65.6} &  \textbf{57.7} & \textbf{53.8} & \textbf{71.9} &  \textbf{45.0} & \textbf{44.0} & \textbf{58.0} &  \textbf{36.4} \\
\bottomrule
\end{tabular}
\vspace{.0em}
\caption{Comparison to imbalanced recognition-inspired methods.}\label{tab:imbalanced}
\vspace{1em}
\end{table*}

In \cref{fig:largeK_analysis_another}, we also show the per-class prediction distributions using different category numbers.
The results on the class-balanced ImageNet-100 are consistent with the results on CIFAR100 and CUB in the main paper, using a loose category number greater than the ground truth may harm fitting the ground-truth class distribution, yet the model still manages to find the ground truth category number. Interestingly, we also find that for the long-tailed Herbarium 19 dataset, using a greater category number could in fact help fit the ground-truth distribution.

\subsection{Relationship to Imbalanced Recognition}
Our work also shares motivation with literature in long-tailed/imbalanced recognition~\cite{menon2020long,wang2020long,ren2020balanced}, in which resolving the imbalance in models' prediction is also an important issue.
Technically, they commonly depend on a prior class distribution to adjust classifiers' output, which is not accessible in GCD since labels for novel classes are unknown. One could also estimate this distribution online from predictions, which is inaccurate due to its open-world nature. We note one baseline (ORCA~\cite{cao21orca}) compared in the paper also shares key intuition with these works (adaptive margin). We also reimplement one close work that operates on imbalanced semi-supervised learning, \ie, DebiasPL~\cite{wang2022debiased}, aligning representation learning with GCD, and show a comparison in \cref{tab:imbalanced}.
DebiasPL surpasses UNO+ on fine-grained classification in novel classes and verifies it could overcome the prediction imbalance to some extent. It also outperforms ORCA but still lags behind GCD and ours. We hypothesise manually altering logits may not be suitable for open-world settings. Instead, a more natural and general solution could be to regularise prediction statistics and let the model adjust via optimisation.

\end{document}